\documentclass{article}

\usepackage{arxiv}

\usepackage[utf8]{inputenc} 
\usepackage[T1]{fontenc}    
\usepackage{hyperref}       
\usepackage{url}            
\usepackage{booktabs}       
\usepackage{amsfonts}       
\usepackage{nicefrac}       
\usepackage{microtype}      
\usepackage{lipsum}		
\usepackage{graphicx}
\usepackage[square,numbers]{natbib}
\usepackage{doi}

\usepackage{array}
\usepackage{pifont}  

\newcommand{\cmark}{\ding{51}} 

\title{A Survey of AI for Materials Science: Foundation Models, LLM Agents, Datasets, and Tools}

\date{} 					

\author{ \hspace{1mm}
	\hspace{1mm}Minh-Hao Van \footnotemark[1] \\
	Department of EECS\\
	University of Arkansas\\
	Fayetteville, AR \\
	\texttt{haovan@uark.edu} \\
    \And
    Prateek Verma \thanks{Both authors contributed equally to this research.} \\
	Department of EECS\\
	University of Arkansas\\
	Fayetteville, AR \\
	\texttt{prateek@uark.edu} \\
    \And
	\hspace{1mm}Chen Zhao \\
	Department of CS\\
	Baylor University\\
	Waco, TX \\
	\texttt{chen\_zhao@baylor.edu} \\
    \And
	\hspace{1mm}Xintao Wu \\
	Department of EECS\\
	University of Arkansas\\
	Fayetteville, AR \\
	\texttt{xintaowu@uark.edu} \\
}

\renewcommand{\shorttitle}{Survey of AI for MS}

\hypersetup{
pdftitle={A Survey of AI for Materials Science: Foundation Models, LLM Agents, Datasets, and Tools},
pdfauthor={Minh-Hao Van, Prateek Verma, Chen Zhao, Xintao Wu},
pdfkeywords={Survey, Materials Science, Foundation Models, LLM Agents, Datasets, Tools},
}

\begin{document}
\maketitle

\begin{abstract}

Foundation models (FMs) are catalyzing a transformative shift in materials science (MatSci) by enabling scalable, general-purpose, and multimodal AI systems for scientific discovery. Unlike traditional machine learning models, which are typically narrow in scope and require task-specific engineering, FMs offer cross-domain generalization and exhibit emergent capabilities. Their versatility is especially well-suited to materials science, where research challenges span diverse data types and scales. This survey provides a comprehensive overview of foundation models, agentic systems, datasets, and computational tools supporting this growing field. We introduce a task-driven taxonomy encompassing six broad application areas: data extraction, interpretation and Q\&A; atomistic simulation; property prediction; materials structure, design and discovery; process planning, discovery, and optimization; and multiscale modeling. We discuss recent advances in both unimodal and multimodal FMs, as well as emerging large language model (LLM) agents. Furthermore, we review standardized datasets, open-source tools, and autonomous experimental platforms that collectively fuel the development and integration of FMs into research workflows. We assess the early successes of foundation models and identify persistent limitations, including challenges in generalizability, interpretability, data imbalance, safety concerns, and limited multimodal fusion. Finally, we articulate future research directions centered on scalable pretraining, continual learning, data governance, and trustworthiness.

\end{abstract}


\section{Introduction}
\label{sec:introduction}

The field of materials science is entering a new era of data-driven discovery, accelerated by advances in artificial intelligence (AI) and machine learning (ML). Traditionally, computational materials science has relied heavily on first-principles simulations such as density functional theory (DFT), molecular dynamics (MD), and finite element methods to predict properties of materials, understand mechanisms, and guide experimental design. However, these methods are computationally intensive, grounded in approximations, and often constrained to small, well-characterized systems. In recent years, machine learning models trained on curated datasets, comprising both simulated and experimental results, have begun to supplement these traditional simulations, enabling faster property prediction and the emergence of generative design capabilities \cite{pyzer2025foundation,wang2024crystalline, han2024ai, ramos2025review, karande2022strategic, mobarak2023scope}. Yet, most of these models remain task-specific, requiring dedicated architectures and training pipelines tailored to each property, material type, or data modality. Their generalization capacity, scalability, and cross-domain adaptability are thus limited.

Inspired by the transformative impact of foundation models (FMs) in natural language processing (NLP) (e.g., BERT \cite{devlin2019bert}, GPT \cite{radford2019language, brown2020language, achiam2023gpt}, PaLM \cite{chowdhery2023palm}) and computer vision (e.g., CLIP \cite{radford2021learning}, DINO \cite{caron2021emerging}), the materials science community is now exploring how similar large-scale and pretrained models might unlock new opportunities for research and innovation. Foundation models are typically defined as large, pretrained models trained on broad, diverse datasets and capable of generalizing across multiple downstream tasks with fine-tuning or prompt engineering. Their hallmark is the emergence of capabilities not explicitly programmed during training and the ability to transfer knowledge across domains, for example, from text to images or from property prediction to generative design, and aid in a variety of downstream tasks.

Materials foundation models aim to inherit these strengths while addressing the unique challenges of the physical sciences. First, materials data is inherently multimodal, comprising structures (atomic, crystalline, polymeric, and multiscale), textual descriptions, experimental data in the form of numbers, tables, and plots, images and spectra, experimental metadata, and simulated predictions or interpolations. Second, material properties emerge from the structure, assembly, and complex interaction of components at a variety of length scales ranging from subatomic, atomic, nanoscopic, microscopic, mesoscopic, and macroscopic. Third, many tasks require strict adherence to physical laws, such as energy conservation and symmetry constraints. Fourth, materials science suffers from limited labeled data; unlike NLP, it lacks billion-scale labeled corpora, relying instead on data that is costly to generate and often imbalanced.

Despite these challenges, recent advances illustrate the promise of this new paradigm. GNoME (Graph Networks for Materials Exploration) discovered over 2.2 million new stable materials by combining graph neural networks with active-learning-driven DFT validation~\cite{merchant2023scaling}. MatterSim, a zero-shot machine-learned interatomic potential (MLIP), is trained on 17 million DFT-labeled structures and supports universal simulation across all elements and a wide range of temperatures and pressures~\cite{yang2024mattersim}. MACE-MP-0, another universal MLIP, achieves state-of-the-art accuracy for periodic systems while preserving equivariant inductive biases~\cite{batatia2023foundation}. Generative approaches such as MatterGen~\cite{zeni2023mattergen}, DiffCSP++~\cite{jiao2024space}, and CrystalFormer~\cite{cao2024space} enable conditional and multi-objective materials generation. Multimodal and cross-domain models like nach0~\cite{livne2024nach0}, MultiMat~\cite{moro2025multimodal}, and MatterChat~\cite{tang2025matterchat} further demonstrate reasoning over complex combinations of structural, textual, and spectral data.

Other efforts aim to build generalist models that unify multiple domains and input types. ATLANTIC~\cite{munikoti2023atlantic} explores cross-modal learning from literature, structures, and properties, while CrystaLLM~\cite{antunes2024crystal} and GT4SD~\cite{manica2023accelerating} provide frameworks for pretraining and multitask evaluation. Autonomous labs such as A-Lab integrate surrogate models and robotic synthesis to optimize experimental discovery~\cite{szymanski2023autonomous}. Process-aware foundation models like Marcato FM~\cite{marcato2024developing} and applications to industrial-scale materials workflows~\cite{ren2025foundation} extend this paradigm to large-scale engineering systems and materials failure prediction.

Moreover, while initial FM research focused on crystalline, inorganic materials, there is growing recognition of the need to represent polymers, soft matter, disordered solids, and biological materials~\cite{choi2025perspective, lee2023towards}. These domains present challenges due to flexible, irregular, or long-range representations, and are motivating the design of new architectures, training data pipelines, and tokenization schemes. Early works such as AtomGPT~\cite{choudhary2024atomgpt} and MoL-MoE~\cite{soares2024multi} are exploring this space.

Large language models (LLMs) are also being adapted for materials science. nach0 unifies natural and chemical language processing and performs tasks like molecule generation, retrosynthesis, and question answering~\cite{livne2024nach0}. Similarly, ChemDFM~\cite{zhao2024chemdfm}, LLaMat~\cite{mishra2024foundational}, and SciTune~\cite{horawalavithana2023scitune} are specialized LLMs trained on scientific literature and domain-specific data. These models enable tasks such as named entity recognition, synthesis extraction, literature summarization, and image-caption alignment. 

LLM agents, which utilize LLMs as core reasoning components and interact with external environments, have been developed to support and automate tasks related to materials science. Recent studies have investigated the development of LLM-based agentic systems for materials science applications \cite{zhang2024honeycomb, jia2024llmatdesign,kang2024chatmof,bazgir2025matagent,ni2024matpilot,m2024augmenting}.  HoneyComb \cite{zhang2024honeycomb} is designed to extend LLM capabilities in the materials science domain. LLMatDesign \cite{jia2024llmatdesign} is proposed to facilitate materials discovery by leveraging state-of-the-art LLMs. ChatMOF \cite{kang2024chatmof} presents an autonomous framework for predicting and generating metal-organic frameworks. MatAgent \cite{bazgir2025matagent} is another LLM-based agentic system tailored for materials science, with a focus on property prediction, hypothesis generation, experimental data analysis, high-performance alloy and polymer discovery, data-driven experimentation, and literature review automation. MatPilot \cite{ni2024matpilot} focuses on literature search, scientific hypothesis generation, experimental scheme design, and autonomous experimental verification, with the goal of developing embodied AI capable of controlling physical robots.

Several toolkits and infrastructure platforms support this growing ecosystem. These include the Open MatSci ML Toolkit~\cite{miret2022open}, designed for standardizing graph-based materials learning workflows, and FORGE~\cite{yin2023forge}, which provides scalable pretraining utilities across scientific domains. Combined, these efforts point to a future of deeply integrated, reusable, and generalizable AI systems for materials science.

\begin{figure}[ht]
    \centering
    \includegraphics[width=1.0\linewidth]{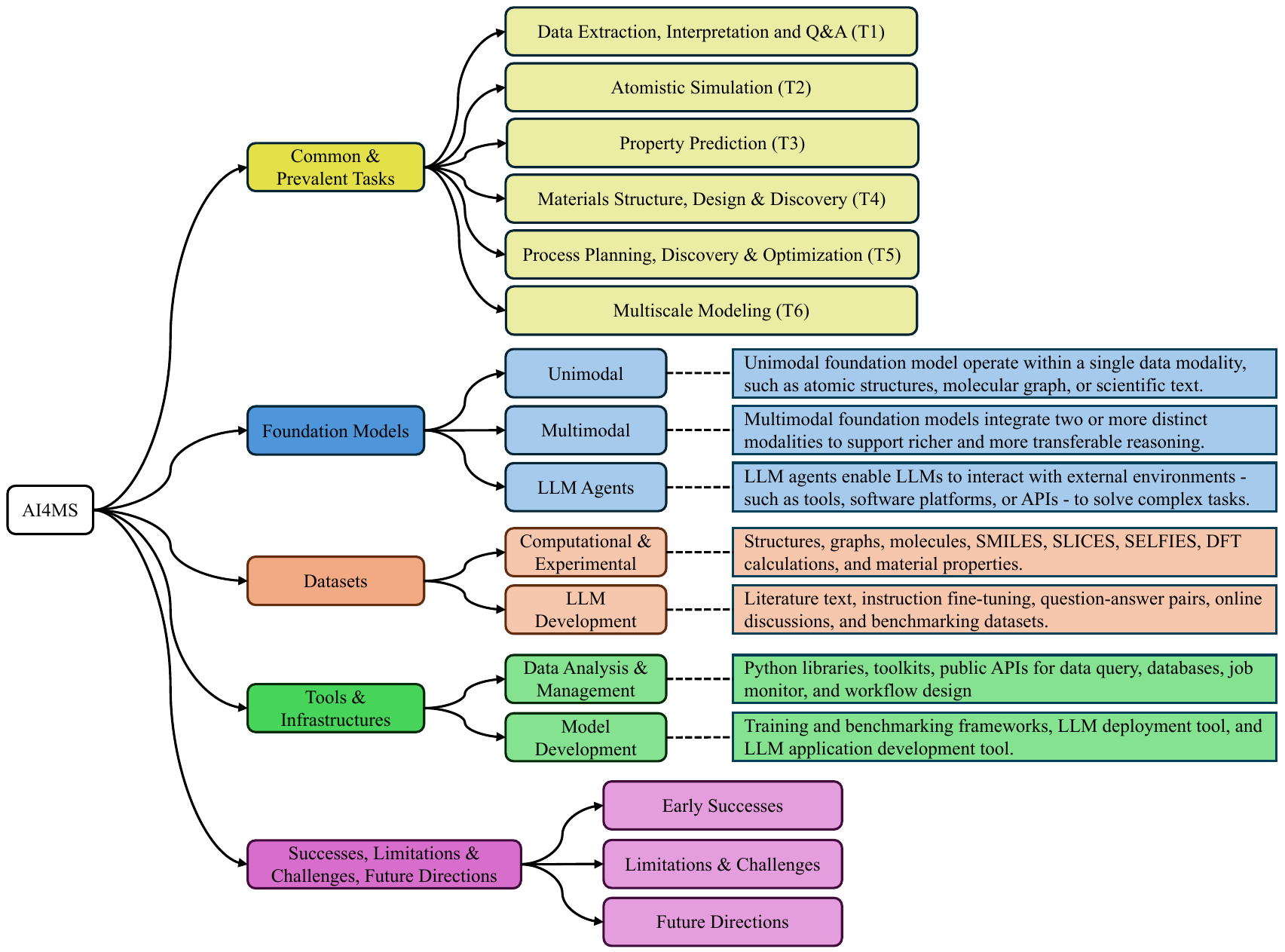}
    \caption{Overview of our survey of AI for materials science (AI4MS), highlighting common tasks, categories of foundation models, datasets, tools and infrastructures, as well as key discussions on early successes, current limitations, challenges, and future directions.}
    \label{fig:ai4ms}
\end{figure}

In this paper, we present a comprehensive survey of foundation models in materials science. We categorize existing models by task, architecture type, and pretraining strategy; highlight multimodal and cross-domain models that bridge structure, text, and property spaces; summarize early successes and their implications for materials discovery pipelines; discuss open challenges such as data bias, long-range interaction modeling, and interpretability; and propose future research directions toward scalable, multimodal, and human-AI collaborative systems. We structure this survey around a proposed taxonomy of foundation models in materials science, spanning six major application areas: data extraction, interpretation and Q\&A; atomistic simulation; property prediction; materials structure, design and discovery; process planning, discovery, and optimization; and multiscale modeling. Figure \ref{fig:ai4ms} shows an overview of our survey of AI for materials science (AI4MS), including common materials science tasks, foundation models, datasets, useful tools and infrastructures for AI materials research, as well as key discussions of successes, limitations, challenges, and future directions. This work aims to serve both as a comprehensive reference and a forward-looking road map for researchers at the intersection of AI and materials science.

In compiling a comprehensive and in-depth survey of foundation models, datasets, and toolkits in materials science, we conducted a systematic literature search using Google Scholar\footnote{Google Scholar, \url{https://scholar.google.com/}}. Our search focused on research related to deep learning, foundation models, LLMs, and LLM-based agentic systems applied to materials science. We further extended the scope to include models in inorganic chemistry, with particular emphasis on crystal and atomic structures, which are highly relevant to materials research. We selectively incorporated high-quality research published in top-tier venues, as well as influential preprints from arXiv\footnote{arXiv, \url{https://arxiv.org/}}. In addition to key papers, we include several open-source toolkits and resources that support the development of foundation models for materials research, selected based on our prior experience and their demonstrated utility in practical applications, as cited in the literature.

Prior to our work, several surveys have addressed the development and application of foundation models in materials science. Notably, Pyzer-Knapp et al. \cite{pyzer2025foundation} focus on foundation models for materials discovery, categorizing them into four primary tasks: data extraction, property prediction, molecular generation, and synthesis prediction. In contrast, our survey encompasses a broader range of tasks relevant to materials science, including atomistic simulations and multiscale modeling. We systematically categorize models into unimodal foundation models, multimodal foundation models, and LLM agents, and further provide a comprehensive overview of available datasets and tools, offering readers additional resources to support and accelerate research in the field. Focusing specifically on crystalline materials, Wang et al. \cite{wang2024crystalline} present a survey of AI-accelerated approaches for crystal discovery, emphasizing four key tasks—property prediction, materials synthesis, characterization assistance, and acceleration of theoretical computations—as well as associated benchmarks, tools, and datasets. By comparison, our survey addresses a wider variety of material classes, including inorganic materials, organic compounds, polymers, and biomaterials, offering a more holistic view of AI applications across the materials science landscape. Another related survey by Han et al. \cite{han2024ai} reviews recent advances in AI-driven inverse materials design, organizing models by material types or model architectures. However, this method of categorization may obscure a clear understanding of AI’s progress in addressing practical, real-world challenges within specific materials science tasks. Expanding into the broader chemistry domain, Ramos et al. \cite{ramos2025review} discuss recent advances in LLMs and agentic AI for chemistry-related tasks. While there is considerable overlap between materials science and chemistry—particularly in tasks such as property prediction, synthesis planning, and information extraction—our survey remains firmly grounded in the core challenges and foundational models specific to materials science.


\section{Common and Prevalent Tasks}
\label{sec:tasks}

In addition to categorizing foundation models by domain or architecture, it is useful to understand the broad functional tasks these models are designed to perform. Below, we organize key AI-driven tasks in materials science into six categories (T1–T6), describing where foundation models offer unique value and how these tasks span across material classes and length scales.

\subsection{Data Extraction, Interpretation and Q\&A (T1)}

A significant portion of scientific knowledge in materials science is locked within unstructured data sources such as research papers, patents, lab notebooks, and experimental reports. The ability to read, interpret, and extract structured information from these documents is foundational to enabling data-centric discovery. Tasks in this area may include document classification, named entity recognition (e.g., identifying materials, properties, synthesis steps), synthesis route extraction, and scientific question answering. Foundation models, particularly large language models (LLMs) and multimodal Transformers, offer a generalizable framework across these varied tasks and data types with minimal retraining~\cite{pyzer2025foundation, mishra2024foundational, choi2025perspective}. They can be tuned or prompted to extract synthesis protocols, material properties, designs and recommendations, or summarize literature trends at scale~\cite{szymanski2023autonomous}. These efforts move beyond information retrieval and aim to construct structured, queryable knowledge graphs from raw scientific content.

\subsection{Atomistic Simulation (T2)}

Atomistic simulation tasks aim to replicate or accelerate quantum and molecular-scale simulations using AI. These include energy and force prediction, structure optimization, and molecular dynamics. Traditionally, each material or system required a custom force field or expensive ab initio calculations. Foundation models trained on millions of DFT-calibrated structures can now serve as general-purpose simulators across diverse chemistries and environments, offering near-DFT accuracy at a fraction of the computational cost~\cite{yang2024mattersim, batatia2023foundation}. These models are especially effective for hard materials and periodic systems. However, they still face limitations in modeling long-range interactions, capturing rare events, and generalizing to non-equilibrium and disordered phases such as liquids, amorphous materials, or multi-component systems.

\subsection{Property Prediction (T3)}

Predicting material properties from composition or structure is one of the most mature and widely adopted tasks in materials science. This includes electronic, mechanical, thermal, optical, and chemical properties across a range of material classes. Traditional models often rely on domain-specific descriptors or task-specific neural networks trained on small datasets. Foundation models shift this paradigm by learning generalizable representations from large, diverse datasets, enabling transfer across tasks and domains~\cite{pyzer2025foundation, choi2025perspective}. Once pretrained, these models can be fine-tuned or applied directly to new prediction tasks with minimal supervision. While most progress has centered on crystalline and small molecular systems, extending these models to disordered materials, porous frameworks, and polymers remains an ongoing challenge. Current models also focus primarily on equilibrium properties, while dynamic or temperature-dependent properties remain less explored~\cite{batatia2023foundation}.

\subsection{Materials Structure, Design, and Discovery (T4)}

Materials design encompasses tasks that involve generating or identifying new candidate materials with specified properties or performance metrics. This includes both molecular and extended solid-state systems, and can be conditioned on targets such as stability, conductivity, reactivity, or processability. Foundation models enable inverse design by learning structure-property relationships in reverse, allowing models to suggest new candidates given a desired property profile. Generative models, including diffusion models, graph-based Transformers, and LLM-driven molecular encoders, now support property-aware molecule and crystal generation, multi-objective optimization, and structure editing~\cite{zeni2023mattergen}. These approaches outperform traditional screening and optimization pipelines in navigating high-dimensional design spaces. However, synthesizability, dynamic stability, and real-world feasibility remain weak points, and polymer or disordered systems introduce additional complexity due to their flexible, long-range, and irregular representations.

\subsection{Process Planning, Discovery, and Optimization (T5)}

The successful realization of new materials requires not only design but also viable synthesis and processing pathways. Process-aware tasks include synthesis planning, reaction condition prediction, experimental design, and decision-making in closed-loop laboratories or production-line industries. Foundation models are increasingly used to extract and learn from synthesis protocols, suggest reaction pathways, and optimize experiment design in autonomous or semi-autonomous labs~\cite{szymanski2023autonomous}. Tasks include recommending precursors, tuning process parameters, or guiding robotic experimentation. This category also encompasses broader workflows such as high-throughput screening, candidate ranking, uncertainty-aware exploration, and multi-property optimization. Foundation models provide scalable and generalizable interfaces for integrating these tasks into adaptive pipelines~\cite{mishra2024foundational}. However, most applications remain constrained to well-studied inorganic materials. Incorporating real-world constraints like cost, scalability, toxicity, and manufacturability into these systems remains an open challenge, especially for polymers, biomaterials, composites, and other complex assemblies.

\subsection{Multiscale Modeling (T6)}

Materials performance often depends not only on atomistic configuration but also on structure and behavior at mesoscopic and macroscopic scales. Multiscale modeling tasks aim to integrate data across these length scales from atomic structure and microstructure to processing conditions, product-level properties, and performance degradation. Foundation models in this domain are still nascent but hold the potential to learn representations that capture temporal evolution, processing-structure-property relationships, and scale-bridging dynamics. Such models could complement finite element methods or serve as surrogates for time-dependent simulations. Current work focuses mostly on atomistic or crystal-level inputs—extending to grain boundaries, sintering behavior, fracture evolution, and large-scale failure remains underexplored~\cite{lee2023towards}. Additionally, many technologically important materials such as polymers, gels, biological materials, and disordered solids do not conform to traditional crystalline representations. These materials present unique modeling challenges due to their flexible, irregular, and sequence-dependent structures. Foundation models could support new representations and generative strategies for these systems, including polymer sequence–property prediction, amorphous structure generation, and solvent-material interaction modeling. Developing domain-appropriate data, architectures, and training strategies for these systems is an important direction for future research.


\section{Foundation Models in Materials Science}
\label{sec:fm_in_ms}

\begin{figure}[ht]
    \centering
    \includegraphics[width=.70\linewidth]{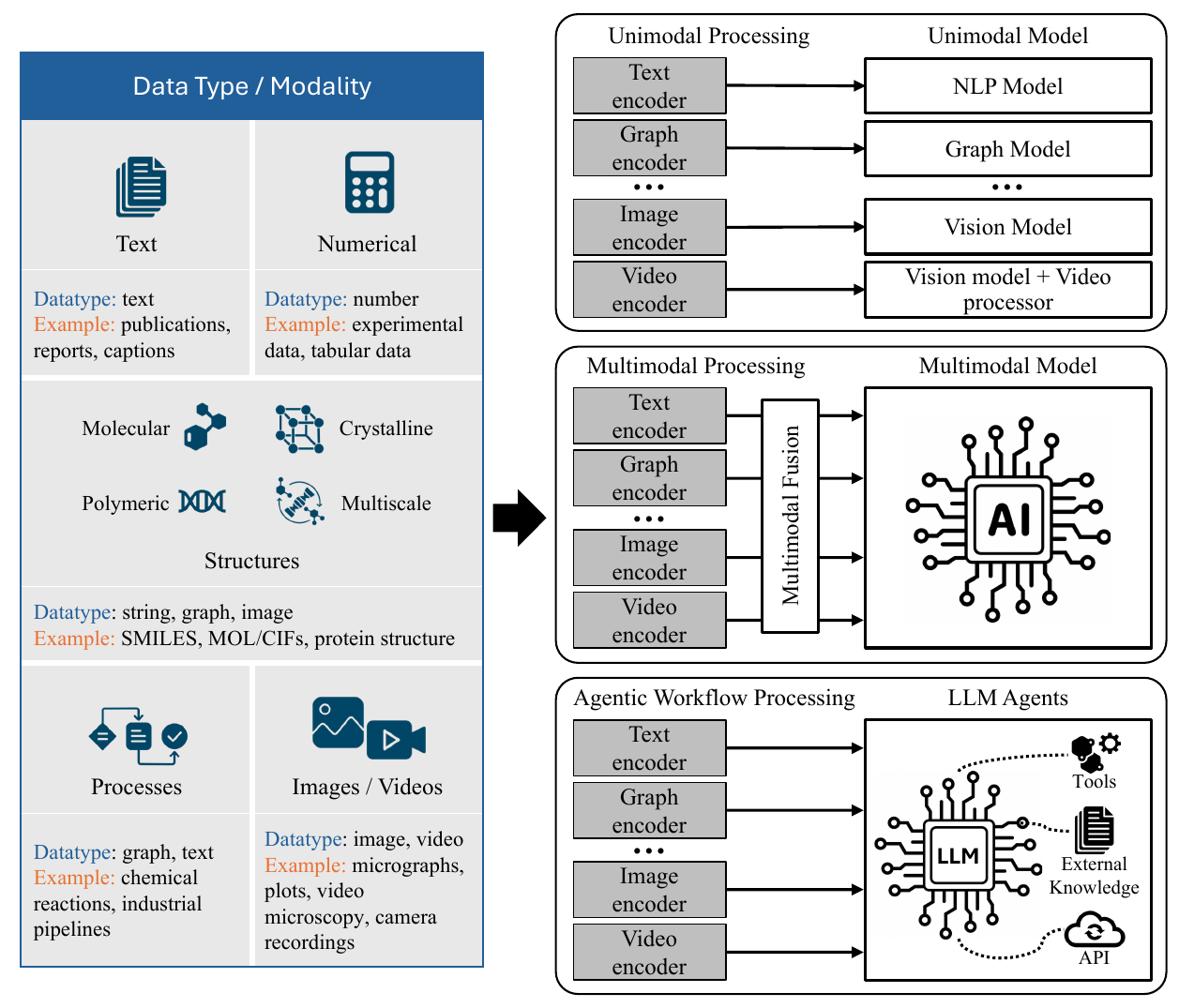}
    \caption{An illustrative example of the interplay of foundation models for materials science with data types and modalities.}
    \label{fig:fm_illustrative_figure}
\end{figure}

Foundation models (FMs) originate in natural language processing (NLP) and demonstrate an effective route to learning generalized representations through the mechanism of self-supervised training on a large corpus of text. In 2017, Vaswani et al. \cite{vaswani2017attention} first introduced the Transformer architecture based entirely on attention mechanisms. Attention operates by comparing each element in an input sequence to every other element. The strength of their association (called weights) captures relationships and contextual information within the sequence. The Transformer architecture is built around two key modules: the encoder and the decoder. The encoder consists of multiple stacked layers, and each layer adopts a self-attention mechanism. The input is tokenized from the model's vocabulary and and each token is represented as a vector that passes through the stacked layers. The decoder is composed of stacked layers, each containing a masked self-attention mechanism that ensures the model only attends to the current and previous tokens, preventing access to future tokens. Moreover, an encoder-decoder attention mechanism is adopted to align the decoder's output with relevant encoder inputs. The decoder can generate the sequence autoregressively, predicting each next token based on prior inputs. The encoder-decoder architecture is ideal for transforming sequences, such as translating from one language to another. The encoder itself can learn generalized representations and thus is ideal for tasks such as property prediction whereas the decoder is ideal for tasks requiring sequence generation or completion such as inferring new outputs from input prompts. 

Drawing from the success of the Transformer architecture, BERT as an encoder-only model, GPT as a decoder-only model, and BART as an encoder-decoder model further demonstrate impressive generalization across tasks such as summarization, translation, and question answering. 
BERT (Bidirectional Encoder Representations from Transformers)~\cite{devlin2019bert} was introduced in 2018 and utilizes only the encoder component. BERT's bidirectional Transformer is pretrained on the unlabeled text and the model processes the context both to the left and right of the word in question, thus enabling BERT to develop more comprehensive representations of input sequences, rather than mapping input sequences to output sequences. GPT (Generative Pretrained Transformer) \cite{radford2018improving} was introduced in 2018 as a decoder-only, left-to-right unidirectional language model to predict the next word in a sequence based on previous words, without an encoder. This decoder-based architecture forms the foundation for state-of-the-art large language models such as GPT-4 \cite{achiam2023gpt}, Google Gemini \cite{team2023gemini}, or Llama \cite{touvron2023llama} models. BART (Bidirectional and Auto-Regressive Transformers) \cite{lewis2019bart}  was introduced in 2019 and is a sequence-to-sequence model consisting of a BERT-like bidirectional encoder and a GPT-like autoregressive decoder. BART is pretrained to reconstruct the original text from the corrupted ones (e.g., deleting tokens and shuffling sentences) with left-to-right autoregressive decoding as in GPT models. All these foundation models are trained on broad corpora using self-supervised objectives and then adapted to specific tasks with minimal fine-tuning. Their scale, in terms of parameters, data, and computing, enables emergent capabilities not explicitly programmed during training. 

Beyond these static foundation models operating solely within their learned parameters and pretraining data, a new paradigm has emerged to support more autonomous and flexible problem-solving: LLM agents \cite{yao2023react,song2023llm,yao2023tree,huang2022language,kosinski2023theory,shanahan2023role,wang2024survey}. These systems enable LLMs to interact with external environments—such as tools, software platforms, external knowledge bases, or APIs—to solve complex tasks. In this paradigm, the LLM serves as a core reasoning agent: understanding the context, analyzing inputs, planning next steps, and taking actions by invoking external tools or resources. LLM agents thus offer a promising framework for interactive scientific discovery, capable of orchestrating design, prediction, simulation, synthesis planning, and information retrieval through prompting, planning, and tool use. These agentic capabilities open up new opportunities for materials science, where complex workflows often require sequential reasoning, hypothesis generation, interaction with simulations, or retrieval from literature text.

The concept of foundation models in materials science is relatively new and continues to evolve. A widely accepted definition describes FMs as large, pretrained models capable of learning transferable representations across diverse modalities and domains that can be reused across a range of downstream applications~\cite{choi2025perspective, pyzer2025foundation}. These models differ from traditional machine learning pipelines, which are typically narrow, task-specific, and require extensive customization for each new task or data modality. In the context of materials science, FMs aim to unify representations across atomic structures, chemical formulas, natural language descriptions, spectroscopic data, and experimental metadata. The goals are to (i) enable transfer learning across tasks and materials systems, (ii) accelerate simulation and design workflows, and (iii) support generalist models for reasoning, generation, and prediction. Some FMs are relatively narrow (e.g., trained only for force prediction or molecular generation), while others aim to span multiple modalities and task types. These are sometimes referred to as "small" and "big" foundation models, respectively~\cite{choi2025perspective}. In this section, we first discuss unimodal foundation models tailored to individual data types, then explore multimodal approaches that integrate diverse inputs, and finally discuss emerging  AI agentic frameworks that aim to assist with complex tasks in materials science. Figure \ref{fig:fm_illustrative_figure} offers an illustrative example of how different types of foundation models process and learn from various data modalities in materials science research. To summarize key developments, Table~\ref{tab:foundation_models} presents a comparative overview of notable foundation models, detailing their names, underlying architectures, supported modalities, datasets used, and associated tasks.

\begin{table}[t]
\centering
\scriptsize
\caption{Representative unimodal, multimodal, and agent-based foundation models in materials science. Each entry includes the model name, underlying architecture, supported modalities, datasets used, and associated AI tasks.}
\begin{tabular}{ >{\raggedright\arraybackslash}p{10em} >{\raggedright\arraybackslash}p{1.3in} >{\raggedright\arraybackslash}p{1.2in} >{\raggedright\arraybackslash}p{1.3in} >{\raggedright\arraybackslash}p{4em}}
\toprule
\textbf{Model Name} & \textbf{Architecture} & \textbf{Modality} & \textbf{Data Sources} & \textbf{Tasks} \\
\midrule
\multicolumn{5}{c}{\textbf{Unimodal Foundation Models}} \\\midrule
HoneyBee \cite{song2023honeybee} & Llama-2 & Instruction fine-tuning (IFT) & MatSci-Instruct & T1 \\
ChemDFM \cite{zhao2024chemdfm} & Llama-2/3  & Literature text & PubChem, Q\&A datasets, textbooks and papers & T1 \\
LLaMat \cite{mishra2024foundational} & Llama-2/3 & Literature text & Papers, RedPajama, MatSci community discourse & T1 \\
LLaMat-Chat \cite{mishra2024foundational} & LLaMat & Instruction fine-tuning (IFT) & OpenOrca, MathQA, MatSciNLP, MatBookQA, MaScQA & T1 \\
MACE-MP-0 \cite{batatia2023foundation} & Equivariant Graph Tensor Network & Atomic graphs & MPTrj & T2, T3 \\
MatterSim \cite{yang2024mattersim} & M3GNet + Graphormer & Atomic structures & MP, Alexandria, self-collected  & T2, T3, T4, T5 \\
CGCNN \cite{xie2018crystal} & GCNN & Crystal structures & MP & T3\\
MEGNet \cite{chen2019graph} & GNN (distinct training for crystals or molecules) & Crystal structures, molecules & MP, QM9 & T3\\
CatBERTa \cite{ock2023catalyst} & RoBERTa & Textual string of structures & OC20 & T3\\
MoLXPT \cite{liu2023molxpt} & GPT-2 & Text, SMILES (text format) & PubMed, PubChem & T3\\
GNoME \cite{merchant2023scaling} & GNN + Ensemble learning & Crystal graphs & MP & T3, T4, T5 \\
CDVAE \cite{xie2021crystal} & VAE & Crystal structures & MP, Perov-5, Carbon-24 & T4 \\
DiffCSP \cite{jiao2023crystal} & Diffusion model & Crystal structures & MP, Perov-4, Carbon-24, MPTS-52  & T4 \\
MatterGen \cite{zeni2023mattergen} & Diffusion model & Crystal structures & MP, Alexandria, ICSD & T4 \\
GP-MoLFormer \cite{ross2024gp} & Transformer & SMILES representations & ZINC, PubChem & T4 \\
CrystaLLM \cite{antunes2024crystal} & GPT-2 & CIFs & MP, OQMD, NOMAD & T4 \\
MatterGPT \cite{chen2024mattergpt} & GPT-2 & SLICES representations & Alexandria & T4 \\
ChemFormer \cite{irwin2022chemformer} & BART & SMILES & ZINC & T5 \\
FlowLLM \cite{sriram2024flowllm} & Riemannian Flow Matching + LLMs & Textual strings of structures  & MP & T4 \\
LLaMat-CIF \cite{mishra2024foundational} & LLaMat & CIF instruction fine-tuning & AMCSD, GNoME, MP & T4 \\
CSLLM \cite{song2024large} & Llama-2/3 & Crystal structures & ICSD, MP, COD, OQMD, JARVIS & T5 \\
\midrule
\multicolumn{5}{c}{\textbf{Multimodal Foundation Models}} \\\midrule
LLM-Fusion \cite{boyar2025llm} & GPT-2 & Text, SMILES, SELFIES, molecular fingerprints & QM9, ChEBI-20 & T3 \\
MoL-MoE \cite{soares2024multi} & Llama-3.2 & SMILES, SELFIES, molecular graphs & ZINC, ChEMBL, Moses & T3 \\
MatterChat \cite{tang2025matterchat} & Material encoder + Bridge model + LLM & Text, atomic structure & MPTrj, MP & T1, T3, T4, T5 \\
Text+Chem T5 \cite{christofidellis2023unifying} & T5 & Text, SMILES & Pistachio, CheBI-20 & T1, T4, T5 \\
nach0 \cite{livne2024nach0} & T5 & Text, SMILES & Pubmed, USPTO, ZINC & T1, T4, T5 \\
SciTune \cite{horawalavithana2023scitune} & LLaVA & Image, text & SciCap, ScienceQA & T1, T4 \\
MultiMat \cite{moro2025multimodal} & CLIP & Crystal structures, density of states, charge densities, text & MP, SNUMAT & T3, T4 \\
Marcato FM \cite{marcato2024developing} & Encoder-Decoder + Llama-3 & Text, simulation grids & Self-collected simulations & T6 \\
\midrule
\multicolumn{5}{c}{\textbf{LLM Agents}} \\\midrule
HoneyComb \cite{zhang2024honeycomb} & BM25 + Contriever + LLM & Literature text, web search & MatSciKB & T1 \\ 
LLMatDesign \cite{jia2024llmatdesign} & MatDeepLearn + TorchMD-Net + LLM & Structure, text & MP & T3, T4 \\ 
ChatMOF \cite{kang2024chatmof} & LLM  as agent + LLM as evaluator + Tools & Structure, text & Collection of MOF databases & T1, T3, T4 \\
MatAgent \cite{bazgir2025matagent} & LLM + ML models for materials science  & Literature text, props & Self-collected data & T1, T3, T4\\ 
MatPilot \cite{ni2024matpilot} & Knowledge retriever + physical workstations + LLM & Tabular data, text, graphs & N/A & T1, T5 \\ 
ChemCrow \cite{m2024augmenting} & CoT reasoning + LLM  & Literature search, web search & Online & T1, T4 \\ 
\bottomrule
\label{tab:foundation_models}
\end{tabular}
\end{table}

\subsection{Unimodal Foundation Models}

Unimodal foundation models operate within a single data modality, such as atomic structures, molecular graphs, or scientific text. These models are typically pretrained on large-scale datasets representative of their domain and often form the backbone of property prediction and simulation workflows.

\subsubsection{Data Extraction, Interpretation and Q\&A (T1)} A significant volume of materials information is contained in documents such as scientific publications, patents, and presentations. Extraction models with information retrieval capability have been extensively applied to identify materials from relevant documents and to link described properties with these materials. 

For materials identification, \textbf{MatBERT} \cite{trewartha2022quantifying} is a BERT-based model trained for materials science, focusing on named entity recognition (NER) to extract and classify entities related to materials science into predefined labels. Being trained on a diverse set of NER datasets, MatBERT achieves impressive performance compared to baselines such as BERT.
\textbf{MatSciBERT} \cite{gupta2022matscibert} is a domain-specific, BERT-based model trained on a large corpus of peer-reviewed materials science publications and establishes state-of-the-art results on downstream tasks such as named entity recognition, relation classification, and abstract classification. 
\textbf{ChemDFM}~\cite{zhao2024chemdfm} is a domain-specific language model trained on chemistry and materials literature. It supports tasks like synthesis information extraction, document classification, and literature-based reasoning.

For property extraction and association, \textbf{MaterialsBERT} \cite{shetty2023general}  builds a general-purpose data extraction pipeline to automatically extract material properties from literature.  MaterialsBERT is trained using 2.4 million materials science abstracts and  obtains $\sim$300,000 material property records that are made available at PolymerScholar\footnote{PolymerScholar, \url{polymerscholar.org}}. Many tools for converting visual representations such as plots and charts into structured tabular data can help enhance  the
overall efficiency and accuracy of data extraction pipelines in materials. For example, \textbf{MolScribe} \cite{qian2023molscribe} is an image-to-graph generation algorithm to identify molecular structures from images in documents. It can predict atoms and bonds, along with their geometric layouts, for the molecular structure construction. \textbf{DePlot} \cite{liu2022deplot} is a tool that translates the image of a plot or chart to a linearized table. The output of DePlot can then be directly used to prompt a pretrained LLM to extract necessary information. 

A separate body of work examines the performance of existing or fine-tuned LLMs on materials science tasks or integrate them with other algorithms for materials science research. 
For example, Zaki et al. \cite{zaki2024mascqa} present a dataset of 650 Q\&A questions and evaluate the performance of Llama-2-70B, GPT-3.5, and GPT-4 models on solving these questions via zero-shot and chain of thought prompting. 
Similarly, Van Herck et al. \cite{van2025assessment} study the performance of fine-tuning three open-source LLMs (GPT-J-6B \cite{wang2021gpt}, Llama3.1-8B \cite{grattafiori2024llama}, and Mistral-7B \cite{jiang2023mistral7b}) for a range of different chemical questions, such as predicting properties of monomers from SMILES (Simplified Molecular Input Line Entry System) \cite{weininger1988smiles}, and benchmark their performance against traditional binary classification models.
\textbf{Uni-SMART} (Universal Science Multimodal Analysis and Research Transformer) \cite{cai2024uni} is designed for understanding multimodal scientific literature that contain molecular structures, chemical reactions, charts, and tables, in addition to textual content. The comparative evaluations demonstrate its superiority over several existing LLMs, such as GPT-4o, Gemini, and Claude \cite{claude}. 
\textbf{HoneyBee} \cite{song2023honeybee} is an early attempt to build LLMs for materials science. A material-focused instruction fine-tuning framework, called \textbf{MatSci-Instruct}, is proposed to generate data for fine-tuning the Llama-based model. MatSci-Instruct utilizes ChatGPT as an instructor model to generate instruction-following data for materials science. Then, Claude serves as a verifier to assess the quality of generated data. Finally, s refinement-feedback loop approach is adopted to train HoneyBee with generated datasets from MatSci-Instruct. Other notable efforts include \textbf{LLaMat} and \textbf{LLaMat-Chat} \cite{mishra2024foundational}, which are based on the Llama-2 and Llama-3 architectures. LLaMat comprises a family of models specialized in general materials science knowledge, distilled from research papers, textbooks, and online forum discussions. LLaMat-Chat is an instruction fine-tuned variant of LLaMat, trained on both general and scientific question-answering datasets, thereby enabling more human-like, interactive dialogue capabilities.

\subsubsection{Atomistic Simulation (T2)}

In the domain of atomic-scale modeling, \textbf{MatterSim}~\cite{yang2024mattersim} and \textbf{MACE-MP-0}~\cite{batatia2023foundation} represent two large-scale efforts to build universal machine-learned interatomic potentials. Trained on tens of millions of DFT-labeled configurations, these models can simulate dynamics, phonon spectra, and phase stability across broad chemical spaces without system-specific retraining. While MatterSim adopts MEGNET \cite{chen2019graph} and Graphormer \cite{ying2021transformers} to build the prediction pipeline, MACE-MP-0 is built on a state-of-the-art architecture, \textbf{MACE} \cite{batatia2022mace}, which utilizes equivariant and many-body message passing. \textbf{ANI} \cite{smith2017ani} utilizes Behler and Parrinello's symmetry functions \cite{behler2007generalized} and neural network potentials to extract molecular representations, which are then utilized to predict molecular energy. Similarly, \textbf{AIMNet} \cite{zubatyuk2019accurate} and \textbf{AIMNet2} \cite{anstine2025aimnet2} introduce atom-in-molecule networks to learn atomic representations and demonstrate their applications in predicting material properties.

\subsubsection{Property Prediction (T3)}

Identifying material properties is a complex task and classical ab
initio physical simulation techniques, such as DFT, are computationally intensive. Deep learning techniques offer an efficient way to predict material properties by capturing the mapping relationship between materials data and properties from the collected datasets. Many of those models can be grouped into GNN-based methods \cite{scarselli2008graph} and Transformer-based methods. Specifically, these models are initially trained to encode the input features of materials (SMILES, SELFIES, crystal structures, etc.) into embedding vectors, which are then used to predict properties. It is important to note that SMILES and SELFIES (Self-Referencing Embedded Strings) \cite{krenn2020self} are 2D molecular representations that omit critical information about a molecule's 3D conformation. Widely used datasets such as ZINC \cite{sterling2015zinc, irwin2020zinc20} and ChEMBL \cite{zdrazil2024chembl, davies2015chembl} include billions of molecules but typically provide only 2D representations. In contrast, datasets of inorganic solids, such as crystalline materials, often include explicit 3D structural information, offering richer features for learning spatially dependent material properties.

Geometric GNNs are models designed to process graph data with geometric information such as spatial coordinates and angles. 
\textbf{SchNet} \cite{schutt2018schnet} is one of the earliest invariant GNN models designed for simulating quantum interactions in molecules using continuous-filter convolutional layers. This method directly models interactions between atoms utilizing distance information, thus achieving rotation invariant energy predictions. \textbf{CGCNN} \cite{xie2018crystal} is designed for handling crystal structures and it can capture long-range interactions and global geometric structures.  This model  represents crystal structures as multi-edge graphs where nodes represent atoms and all their copies within the 3D space, and edges represent the connections between these atoms. CGCNN uses convolution and pooling layers to extract and learn both local and global features of the structure. CGCNN can extract the contributions from local chemical environments to global properties and thus is interpretable.  \textbf{MEGNet} \cite{chen2019graph} is another GNN-based model and is trained on crystal structures from the Materials Project \cite{jain2013commentary} or QM9 \cite{ruddigkeit2012enumeration,ramakrishnan2014quantum} molecules, aiming to predict the formation energies, band gaps, or elastic moduli of crystals. Atomic attributes, bond attributes, and global state attributes serve as input features for MEGNet. In the pipeline, bond attributes are updated first, then atomic attributes, and lastly global state ones. A new graph representation is created from the MEGNet block. Additional set2set and dense layers are then added to predict the final output.  We refer interested readers to a recent survey \cite{wieder2020compact}  that covers 80 GNNs used for property prediction.

Encoder-only Transformer architectures are primarily composed of an encoder, making them well-suited for property prediction  that requires extracting meaningful information from input sequences such as SMILES. 
Encoder-only models based on the  BERT architecture are predominantly used for property prediction as these models would ideally convert SMILES strings into a vector representation, which captures  material properties. BERT-based models for property prediction include 
\textbf{ChemBERTa} \cite{chithrananda2020chemberta}, \textbf{CatBERTa} \cite{ock2023catalyst}, \textbf{SolvBERT} \cite{ock2023catalyst}, and \textbf{Mol-BERT} \cite{li2021mol}. 
\textbf{ChemBERTa} \cite{chithrananda2020chemberta} adapts RoBERTa \cite{liu2019roberta} architecture to property prediction task using SMILES data from PubChem \cite{kim2025pubchem}. \textbf{CatBERTa} \cite{ock2023catalyst} is trained on a set of DFT calculations from Open Catalyst 2020 (OC20) \cite{chanussot2021open} dataset. \textbf{SolvBERT} \cite{ock2023catalyst} is a model trained using an unsupervised scheme on solvation data. To effectively learn molecular representation, \textbf{SELFormer} \cite{yuksel2023selformer} utilizes SELFIES data to train a Transformer-based RoBERTa model for extracting high-quality embeddings. Similarly, \textbf{Mol-BERT} focuses on learning molecular representation using SMILES from ZINC and ChEMBL datasets.

A line of research that utilizes GPT models have been introduced recently for predicting material property as Transformer-based architectures have shown remarkable capability in graph learning. \textbf{Matformer} \cite{yan2022periodic} leverages the geometric distances between atoms from two adjacent unit cells to encode periodic patterns, thus  enabling Matformer to encapsulate the lattice information and periodic patterns.  \textbf{ComFormer} \cite{yan2024complete} converts both  invariant graph representation and equivariant graph representation into embeddings via Transformers, enabling ComFormer to capture both local and global geometric information of different structures. \textbf{SMILES-GPT} \cite{adilov2021generative} is a pretrained GPT-2-based language model on a large SMILES corpus from PubChem. \textbf{MolXPT} \cite{liu2023molxpt} is introduced as a GPT-2-based language model using both text and SMILES from PubMed \cite{pubmed} and PubChem. \textbf{SPT} \cite{winter2022smile} takes SMILES as input to predict binary limiting activity coefficients, utilizing GPT-3 architecture. Another attempt to fine-tune GPT-3 on chemical and material data is introduced in \cite{jablonka2024leveraging}.

\subsubsection{Materials Structure, Design, and Discovery (T4)} 
This challenging task requires the development of advanced AI models to learn complex material structures and synthesize novel materials. Based on data representation, these methods can be grouped into two categories: geometric graph-based generation and string-based generation.

In the early stage, many GNN-based architectures have been proposed for this task.
G-SchNet \cite{gebauer2019symmetry} is a pioneer in molecule generation. It incorporates the constraints of Euclidean space and the rotational invariances of the atom distribution as prior knowledge. It constructs an equivariant conditional probability distribution to determine the next atomic position by using the distance between the previously placed positions and the next atomic position as constraints. 
\textbf{CDVAE} \cite{xie2021crystal} is proposed to generate stable crystalline materials from known materials, focusing on a generative approach using Diffusion \cite{ho2020denoising} and VAE \cite{kingma2013auto,kingma2019introduction}. The model includes three main components: (1) a periodic GNN encoder, (2) a property predictor, and (3) a periodic GNN decoder, which are optimized concurrently with stable materials. An evaluation of three tasks (reconstruction, generation, and property optimization) demonstrates the capability of CDVAE in generating materials given defined properties. Based on CDVAE, \textbf{Con-CDVEA} \cite{ye2024cdvae} generates
crystals’ latent variables according to given properties such as formation energy or band gap, and then yields the corresponding crystal structure by decoding the latent variables.
\textbf{DiffCSP} \cite{jiao2023crystal} utilizes the fractional coordinate system to intrinsically represent crystals and model periodicity. By employing an equivariant graph neural network for the denoising process, DiffCSP  conducts joint diffusion on lattices and fractional coordinates to capture the crystal geometry, thereby enhancing the modeling of the crystal geometry. DiffCSP separately
and simultaneously adds noise and denoises on the fractional coordinates, atom types, and lattices. For the generation stage, random noises are sampled from Gaussian space for denoising as new materials.  \textbf{DiffCSP++}  \cite{jiao2024space}  uses the diffusion model to generate new materials by incorporating the space group constraint. DiffCSP++ first encodes the lattice parameters as invariant vectors to ensure the lattice parameters satisfy the space group constraints. Then, DiffCSP++ generates the atom that satisfies the symmetry constrained by the space group. 

\textbf{MatterGen} \cite{zeni2023mattergen} is proposed to generate stable, diverse inorganic materials across the periodic table. MatterGen uses a diffusion model to produce crystalline structures by gradually refining atom types, coordinates, and the periodic lattice. Specifically, it uses a large dataset of stable material structures to train an equivariant score network and then fine-tunes the score network with a labeled dataset, where the property labels are encoded to steer the generation under specified property constraints.  
\textbf{GNoME}~\cite{merchant2023scaling} is a graph neural network-based discovery engine and applies large ensembles and uncertainty quantification to predict material stability. It facilitates the discovery of new materials that extend beyond existing data distributions and has successfully discovered over 2 million new candidate crystals through active learning. The GNoME framework consists of  two key modules: symmetry-aware partial substitutions combined with random structure search, and GNN-based modeling of material properties, and drive two independent materials discovery pipelines: structural pipeline and  compositional pipeline. The former focuses on evaluating the stability of crystal frameworks without considering specific atomic types, filtering randomly generated structures using GNoME to retain potentially stable frameworks whereas the latter takes chemical formulas as input to GNoME, identifying stable chemical combinations to explore novel material compositions. DFT calculations are then performed to further validate their structural stability. Stable materials are  added to the training set for subsequent iterations, creating an iterative active learning loop.

Crystalline materials are stored in standard text file formats known as CIF (Crystallographic Information File) or SLICES (Simplified Line-Input Crystal-Encoding System \cite{xiao2023invertible}). By treating CIFs or SLICES as plain string representations, many works explore the use of generative language models to generate crystals. These works \cite{antunes2024crystal, gruver2024fine, flam2023language} are different from models that use graph and graph-derived string representations as they  treat materials as a sequence of discretized tokens and adopt Transformer architecture, thus utilizing the capability of next-token prediction given the input sequences for material generation. 
A majority of works for material structural generation are based on decoder-only GPT models. GPT employs positional encodings to maintain word order in its predictions. Its self-attention mechanism prevents tokens from attending to future tokens, ensuring each word prediction depends only on preceding words. \textbf{CrystaLLM} \cite{antunes2024crystal} is a decoder-only Transformer-based tool for crystal structure generation trained on an extensive corpus of the CIFs representing the structures of millions of inorganic solid-state materials. During training, the model is given a sequence of tokens from the corpus of CIFs, and is tasked with predicting the tokens that follow each of the given tokens. After training, the model can be used to generate new CIFs, conditioned on some starting sequence of tokens. 
For molecular design, \textbf{GP-MoLFormer}~\cite{ross2024gp} enables property-conditioned molecule generation using Transformer architectures trained on large molecular datasets. 

\textbf{CrystalFormer} \cite{cao2024space} is a Transformer-based autoregressive
model specifically designed to generate crystal materials that respect space group symmetries. It enriches the edge features' expressiveness by further incorporating angular information into the edge features and  further introduces a graph construction method specifically designed for periodic invariance.
\textbf{MatterGPT} \cite{chen2024mattergpt} is another Transformer-based model trained for generating solid-state materials with given properties. Different from other language-based models which usually receive textual input, MatterGPT receives SLICES representation as input features and then generates output SLICES of expected materials. Flam-Shepherd et al. \cite{flam2023language} demonstrate that language models trained directly on sequences derived directly from chemical file formats like XYZ files, CIFs, or Protein Data Bank (PDB) files can directly generate molecules, crystals, and protein binding sites in three dimensions. Gruver et al. \cite{gruver2024fine} show that fine-tuned LLMs can generate the three-dimensional structure of stable crystals as text. 

Being trained on a vast corpus of data, LLMs demonstrate impressive performance in multiple downstream tasks. \textbf{LLaMat-CIF} \cite{mishra2024foundational} is an instruction fine-tuned model with CIF data based on the pretrained LLaMat model. The goal is to enable the capability of LLMs in understanding and generating useful information with CIFs as inputs.  
\textbf{AtomGPT} \cite{choudhary2024atomgpt} leverages GPT-2 and quantized Mistral models to learn the complex relationships between atomic structures and material properties from datasets such as JARVIS-DFT \cite{choudhary2020jarvis} and supports both property prediction and structure generation. While LLaMat-CIF and AtomGPT aim to train customized LLMs for materials science tasks, other works utilize strong pretrained LLMs to solve these tasks. \textbf{MatLLMSearch} \cite{gan2025large} integrates pretrained LLMs with evolutionary search algorithms and supports crystal structure generation, crystal structure prediction, and multi-objective optimization of properties, all without fine-tuning. Beyond the direct use of LLMs to generate crystal representations, recent approaches have explored integrating LLMs with other generative techniques to enhance model performance. \textbf{FlowLLM} \cite{sriram2024flowllm} combines LLMs with Riemann Flow Matching (RFM) to design novel crystalline materials.  Specifically, FlowLLM first fine-tunes an LLM to learn a base distribution of meta-stable crystals in a text representation.  These text representations are then converted into geometric graph representations. The RFM model takes samples from the LLM and iteratively refines the atom coordinates and lattice parameters to produce stable crystal structures. FlowLLM is trained on the widely used dataset of inorganic crystalline materials, derived from the Materials Project, focusing on a subset of compounds with up to 20 atoms known to be metastable.

\subsubsection{Process Planning, Discovery, and Optimization (T5)}

Synthesis planning aims for experimental realizations, i.e., predicting synthesizability and proposing the right precursors, pathways, and conditions to synthesize the targeted materials. \textbf{Molecular Transformer} \cite{schwaller2019molecular} first applies a Transformer for synthesis prediction, in particular, translating reactants and reagents into the final product.  This work studies the correlations between chemical motifs in reactants, reagents, and products in the USPTO dataset \cite{uspto}.

Advancements in synthesis prediction are mainly based on the BART encoder-decoder architecture \cite{lewis2019bart}. \textbf{ChemFormer} \cite{irwin2022chemformer} shows that models pretrained using only the encoder stack are limited for sequence-to-sequence tasks.  ChemFormer is based on the BART encoder-decoder architecture and is trained with 100M SMILES from ZINC. The evaluations on datasets such as ChEMBL and ESOL \cite{delaney2004esol} demonstrate state-of-the-art results in both sequence-to-sequence synthesis tasks and discriminative tasks. 

Recent approaches have explored the use of LLMs for synthesis prediction. \textbf{MatChat} \cite{chen2023matchat} demonstrates the effectiveness of LLMs in predicting the synthesizability of inorganic compounds and selecting suitable precursors. 
\textbf{CSLLM} \cite{song2024large} comprises three LLMs designed to predict material synthesizability, synthesis methods, and synthesis precursors.  \cite{song2024large} employs Llama-7B fine-tuned via LoRA \cite{hu2022lora} and utilizes a proprietary Materials String representation to encode  crystal structures. It further creates a synthesizability dataset containing 140,120 crystal structures.  
\textbf{SynAsk} \cite{zhang2025synask} is an organic chemistry domain-specific LLM platform.  SynAsk fine-tunes an LLM with domain-specific data and integrates it with a chain-of-thought approach. It supports functionalities such as molecular information retrieval, reaction performance prediction, and retrosynthesis prediction. 
\textbf{MatSci-LLM} \cite{miret2024llms} presents an LLM-based framework for generating experimental hypotheses of real-world materials discovery. By incorporating large-scale multimodal datasets for materials knowledge, MatSci-LLM framework can enable general-purpose LLMs to execute the materials discovery task via a six-step process: (1) materials query, (2) data retrieval, (3) materials design, (4) insilico evaluation, (5) experiment planning, and (6) experiment execution.

\subsubsection{Multiscale Modeling (T6)}

Materials simulations span length scales from atomic to macroscopic and time scales from femtoseconds to hours. Machine learning or deep learning based multi-scale modeling and fusion methods for materials research have been developed \cite{fish2021mesoscopic}. For example, \textbf{MuMMI} \cite{ingolfsson2023machine} is an ensemble machine learning approach for solving protein-membrane interactions that require multiscale modeling. This method connects three resolution scales: (1) coarsest scale (1000 nm), (2) coarse-grained scale (30 nm -- 140K particles), and (3) all-atom scale (30 nm -- 1.4M particles). CNN-based model is utilized by \cite{aldakheel2023efficient} to achieve efficient multiscale modeling of heterogeneous materials, focusing on the prediction of homogenized macroscopic stress given a microstructure input image.
However, there are no existing foundation model architectures specifically designed for multiscale modeling problems in materials science. This is partially because multiscale datasets are not yet well-established. 
Developing foundation models to establish connections and couplings across scales in materials remains an open challenge.

\subsection{Multimodal Foundation Models}

Multimodal foundation models integrate two or more distinct modalities including structure, text, spectra, and images to support richer and more transferable reasoning. These models are especially well-suited for capturing the interconnected nature of materials data, where a material may be simultaneously described by its atomic configuration, synthesis process, performance measurements, and visual or spectroscopic features.

\textbf{Text+Chem T5} \cite{christofidellis2023unifying} is a multi-domain,
multi-task Transformer based language model and provides a unified representation between
natural language and chemical representations. It is designed to support both mono-domain tasks and cross-domain tasks. The mono-domain tasks include molecule-to-molecule where the model predicts the outcome of a chemical reaction based on the starting chemicals or the starting chemicals that would be required to synthesize a given compound, and text-to-text where the model generates the action sequence based on a certain chemical reaction described in natural language. The cross-domain tasks include text-to-molecule where the model takes a textual description of a molecule as an input and generates its SMILES representation, and molecule-to-text where the model takes a molecule represented as SMILES and generates its human-readable textual description. The authors use multiple datasets including Pistachio \cite{mayfield2017pistachio} and CheBI-20 \cite{edwards2021text2mol} in their evaluation. Another example is \textbf{nach0}~\cite{livne2024nach0}, a multitask Transformer that unifies SMILES strings and natural language, enabling molecule generation, retrosynthesis, reaction prediction, and scientific question answering in a single model. \textbf{ATLANTIC}~\cite{munikoti2023atlantic} extends this idea by aligning graph-based chemical representations with textual data, supporting synthesis reasoning and property prediction through interdisciplinary learning. \textbf{Regression Transformer} \cite{born2023regression} abstracts regression as a conditional sequence modeling problem and bridges sequence regression and conditional sequence generation by using a nominal-scale training objective on combinations of numerical and textual tokens. It supports tasks of  property prediction and conditional molecular design. 
Regression Transformer uses MoleculeNet \cite{wu2018moleculenet} and TAPE (Tasks Assessing Protein Embeddings) \cite{rao2019evaluating} benchmarks in their evaluation.

\textbf{LLM-Fusion} \cite{boyar2025llm}  is a multimodal fusion model that leverages LLMs to integrate diverse representations, such as
SMILES, SELFIES, text descriptions, and molecular fingerprints,
for property prediction. Those diverse modalities are embedded and fused into a unified representation. In particular, the encoder for each modality can be of any architecture and can be frozen or fine-tuned. The encoded vectors via projection layers are stacked and enriched with positional encodings.  The unified representation is fed at the input embeddings layer of the LLM, thus skipping the tokenization and positional encoding addition steps of transfomer training. It uses the MoleculeNet-QM9 dataset and ChEBI-20 dataset. \textbf{MoL-MoE} \cite{soares2024multi} is introduced as a Multi-view Mixture-of-Experts framework to predict molecular properties by integrating latent spaces derived from SMILES, SELFIES, and molecular graphs. Mixture-of-Experts (MoE) has become essential for scaling large models by selectively activating sub-networks of experts through a gating network, thereby optimizing training efficiency. MoL-MoE utilizes SMI-TED (289M) foundation model \cite{soares2024large} as the SMILES encoder, SELFIES-BART as the SELFIES encoder \cite{priyadarsini2024self},  and MHG-GNN \cite{kishimoto2023mhg} as the graph encoder. 
A set of nine distinct benchmark datasets sourced from MoleculeNet are used for both classification and regression tasks. \textbf{SciTune}~\cite{horawalavithana2023scitune} adopts a vision-language pretraining strategy, similar to CLIP \cite{radford2021learning}, to align images (e.g., figures, microscopy) with captions and scientific questions. It supports multimodal tasks such as caption generation, figure interpretation, and visual question answering. Similarly, \textbf{MultiMat}~\cite{moro2025multimodal} is a contrastive learning framework that learns embeddings across crystal structures, density of states (DOS), charge densities, and natural language labels.

\textbf{MatterChat} \cite{tang2025matterchat} is a versatile structure-aware multi-modal LLM that unifies material structural data and textual user queries and employs a bridging module to align a pretrained universal machine learning interatomic potential with a pretrained LLM. It supports text-based generation for tasks such as  material property prediction, structural analysis, and descriptive language generation. MatterChat consists of three components: the Material Processing Branch for extracting atomic-level embeddings from material structural graphs, the Bridge Model for producing language model-compatible embeddings, and the Language Processing Branch for processing the user's text-based prompt into the language embeddings. Both language embeddings and query embeddings are fed into the LLM to produce the final text output. 
Finally, \textbf{Marcato FM}~\cite{marcato2024developing} is a cross-modal encoder-decoder model for material failure prediction, combining simulation outputs, structure, and grid-based fields for robust forecasting of fracture behavior. 

Together, these models demonstrate the potential of foundation models to act as unified engines for materials understanding, integrating simulation, design, language, and experimentation within a single learning framework. However, challenges remain in aligning modalities, addressing scale disparities, and ensuring physical fidelity across representation spaces.

\subsection{LLM Agents}
LLM agents can be designed to support and automate related tasks in materials science, including generating experimental plans, calling simulation tools, performing evaluation of outputs, and especially optimizing and repeating the experiments based on the outcome of previous experiment cycles.

\textbf{HoneyComb} \cite{zhang2024honeycomb} provides pretrained LLM capabilities to retrieve and analyze comprehensive knowledge in the materials science domain, enhancing the quality of generated content. HoneyComb consists of three main components: MatSciKB, a knowledge base aggregating diverse sources of materials-related information (e.g., arXiv articles, Wikipedia, datasets, textbooks); ToolHub, a suite of tools for retrieving up-to-date information; and Retriever, which extracts relevant knowledge from both MatSciKB and ToolHub using BM25 \cite{trotman2014improvements} and Contriever \cite{izacard2021unsupervised}. This agentic system can be integrated with different LLMs to enhance their domain-specific reasoning capabilities. 
\textbf{LLMatDesign} \cite{jia2024llmatdesign} employs a step-by-step, self-reflective design to accomplish the materials discovery task: (1) receiving human input about the chemical composition and the target property; (2) recommending the modification (i.e., exchange, addition, substitution, or removal); (3) employing machine learning tools such as MatDeepLearn \cite{fung2021benchmarking} and TorchMD-Net \cite{tholke2022torchmd} for property prediction; (4) evaluating the outcome; and (5) repeating the process with alternative modifications if the target property is not achieved. LLMatDesign translates human instructions into appropriate Materials Project API calls, supports material modifications, and evaluates outcomes using provided tools. Prompt engineering is used to guide pretrained LLMs in carrying out each step. This model randomly chooses ten starting materials from the Materials Project and focuses on designing materials with target properties of band gap and formation energy per atom, enabling a more efficient and autonomous approach to materials discovery. 
\textbf{ChatMOF} \cite{kang2024chatmof} is another agentic framework built to predict and generate metal-organic frameworks by leveraging pretrained LLMs, e.g., GPT-4, GPT-3.5-turbo, and GPT-3.5-turbo-16k, to extract key details from textual inputs and deliver appropriate responses.  The system is comprised of three core components (i.e., an agent, a toolkit, and an evaluator) and it supports tasks of data retrieval, property prediction, and structure generation.
\textbf{MatAgent} \cite{bazgir2025matagent} aims to cover a sufficient number of tasks in materials science, such as property prediction, hypothesis generation, experimental data analysis, high-performance alloy and polymer discovery, data-driven experimentation, and literature review automation. Unlike other systems such as HoneyComb and LLMatDesign, MatAgent involves a human-in-the-loop (HITL) approach, allowing for human oversight at key stages to ensure the quality of generated content. The workflow consists of: (1) hypothesis generation and initial review with HITL; (2) central processing, including code generation, data visualization, report writing, and web search for prediction and analysis; (3) quality review to check for non-sense outcome with HITL and move back to previous step if it happens; (4) final human review; and (5) integration with external tools or databases. 
Similar to MatAgent, \textbf{MatPilot} \cite{ni2024matpilot} also employs human-in-the-loop approach, with a particular emphasis on literature search, scientific hypothesis generation, experimental scheme design, and autonomous experimental verification. MatPilot is designed with two primary modules: (1) a cognition module, which retrieves information from a knowledge base and generates experimental schemes under the guidance of human experts; and (2) an execution module, which autonomously performs experimental tasks based on the plan produced by the cognition module. Unlike previous agents designed to interact with tools or APIs, MatPilot moves toward interaction with physical robots, aiming to establish embodied AI capable of performing real-world experimental procedures. 
\textbf{ChemCrow} \cite{m2024augmenting} aims to build an LLM agent for chemistry, covering several tasks including organic synthesis, drug discovery, and materials design implemented with the LangChain framework. This LLM-based agent utilizes a chain-of-thought reasoning loop to (1) create plans, (2) select appropriate tools, (3) take actions, and (4) analyze the outputs. Incorporating a set of 18 chemistry tools covering general tools, molecule tools, safety tools, chemical reaction tools, ChemCrow enables LLM with autonomous experimentation in chemistry tasks.



\section{Datasets, Tools, and Infrastructures}
\label{sec:infrastructure_datasets}

\subsection{Datasets}
The effectiveness of foundation models in materials science hinges critically on the availability of high-quality, large-scale datasets that span diverse materials classes, properties, and modalities. Over the past few years, a growing number of datasets have emerged to support pretraining and evaluation, many of which are summarized in Table~\ref{tab:datasets}. These datasets vary widely in terms of modality - ranging from atomic structures, compositions, and energy landscapes to text, graphs, and synthesis protocols. 

\begin{table}[ht]
\centering
\scriptsize
\caption{Representative datasets used for training and evaluating foundation models in materials science (MatSci). The table covers both computational/experimental and LLM-oriented datasets, detailing their modalities, sample size, and relevance to key AI tasks (T1–T6).}
\begin{tabular}{ >{\raggedright\arraybackslash}p{3.5cm} >{\raggedright\arraybackslash}p{3.0cm} >{\raggedright\arraybackslash}p{3.0cm} >{\raggedright\arraybackslash}p{1.5cm} > {\raggedright\arraybackslash}p{1.5cm}}
\toprule
\textbf{Dataset} & \textbf{Data Description} & \textbf{Modalities} & \textbf{\# Samples}  & \textbf{Tasks} \\
\midrule
\multicolumn{5}{c}{\textbf{Computational and experimental datasets}} \\\midrule
ICSD \cite{zagorac2019recent} & Inorganic materials & Crystal structures, props & $\sim$300k  & T3, T4 \\
Materials Project (MP) \cite{jain2013commentary} & Structure, property & Structured text, graphs & $\sim$200k  & T2, T3, T5, T6 \\
OQMD \cite{kirklin2015open,saal2013materials} & Inorganic crystals & Structure, DFT props & $\sim$1M & T2, T3 \\
NOMAD \cite{draxl2019nomad} & Hypothetical crystals & Crystal structures, props & $\sim$214k  & T3, T4 \\
OMat24 \cite{barroso2024open} & Inorganic materials & Crystal structures, props & $\sim$118M  & T3, T4 \\
SNUMAT \cite{snumat} & Synthesized materials & Crystal structures, DFT props & $\sim$10k  & T3, T4 \\
MPTrj \cite{deng2023chgnet} & Trajectory data & Atomic structures, props & $\sim$1.5M &  T2, T3 \\
Open Catalyst 2020 (OC20) \cite{chanussot2021open} & Catalysis structures, forces & Graphs, 3D coords & $\sim$1.3M  & T2, T3, T5 \\
Alexandria \cite{schmidt2024improving} & 1D, 2D, 3D materials & Molecular structures, DFT calculations & $\sim$5M & T2, T3, T4 \\
QM9 \cite{ruddigkeit2012enumeration,ramakrishnan2014quantum} & Molecules, DFT & SMILES, 3D structures & 134k  & T2, T3 \\
Guacamol \cite{brown2019guacamol} & Benchmark for molecular design & Molecules & $\sim$1.5M & T4\\
Moses \cite{polykovskiy2020molecular} & Benchmark for molecular design & Molecules & $\sim$4.5M & T4\\
ZINC \cite{sterling2015zinc,irwin2020zinc20} & Molecular library & SMILES, conformers & $>$1B  & T3, T4 \\
ChEMBL \cite{zdrazil2024chembl, davies2015chembl} & Bioactive molecules & Molecules & $\sim$2.2M & T3, T4 \\
\midrule
\multicolumn{5}{c}{\textbf{LLM development datasets}} \\\midrule
MatScholar \cite{tshitoyan2019unsupervised,weston2019named} & Literature text & Paper abstracts & $>$5M  & T1 \\
MatSciKB \cite{zhang2024honeycomb} & Knowledge base & Text & $\sim$38k & T1 \\
PubChem \cite{kim2025pubchem} & Proteins, genes, chemical structures, literature text & SMILES, crystal structures, papers, patterns & $>$200M & T1, T3, T4 \\
MatbookQA \cite{mishra2024foundational} & Q\&A from MatSci books & IFT, Q\&A & $\sim$2k & T1 \\
MaScQA \cite{zaki2024mascqa} & Q\&A from engineering exams & IFT, Q\&A & $\sim$1.5k & T1 \\
MatSci-Instruct \cite{song2023honeybee} & MatSci IFT data & IFT, Q\&A & $\sim$52k & T1 \\
MatSciNLP \cite{song2023matsci} & NLP benchmark & Text, Q\&A &  $\sim$170k &  T1, T4, T5 \\
LLM4Mat-Bench \cite{niyongabo2024llm4mat} & LLM benchmark for property prediction & Crystal composition, CIFs, textual descriptions & $\sim$1.9M & T3 \\
LLM4Mol \cite{zhong2024benchmarking} & LLM benchmark for molecular prediction & Molecules, props & $\sim$40k & T4 \\
MACBENCH \cite{alampara2024macbench} & LLM multimodal benchmark for chemistry and MatSci & Q\&A & 628 & T1 \\
\bottomrule
\end{tabular}
\label{tab:datasets}
\end{table}

\subsubsection{Computational and experimental datasets}
In the last two decades, several computational and experimental datasets have been released to support materials science, which are later used to train deep learning and foundation models to further enhance materials exploration and analysis. In the early stage, \textbf{ICSD} \cite{zagorac2019recent} received its first record back in 1913 and was later available on the web in 2003. It is the largest database of inorganic crystal structures, comprising about 300,000 entries. Each entry includes detailed crystallographic information, such as unit cell parameters, space group, atomic coordinates, site occupation factors, molecular formulas, and molecular weights. After that, other common datasets for crystalline solids, material properties, and DFT calculations—such as \textbf{Materials Project (MP)}~\cite{jain2013commentary}, \textbf{OQMD}~\cite{saal2013materials,kirklin2015open}, \textbf{NOMAD} \cite{draxl2019nomad}, \textbf{Open Material 2024 (OMat24)} \cite{barroso2024open}, and \textbf{SNUMAT} \cite{snumat}—have been released to mainly serve as training data for deep learning and foundation models. Specifically, MP provides an open database of over 200,000 materials through web interfaces and APIs. OQMD is another dataset with high-throughput DFT calculations of thermodynamic and structural properties of 1,317,811 materials. NOMAD serves as an open-access platform for managing, analyzing, and sharing materials science data, currently hosting approximately 214,000 structures.
Recently, OMat24, released by FAIR, presents a large-scale database of inorganic materials accompanied by a set of pretrained models. It contains around 118 million structures labeled with properties such as total energy, atomic forces, and cell stress. To generate these crystal structures, the process begins with a random sampling of relaxed structures, followed by one of three refinement techniques: Rattled Boltzmann sampling, ab initio molecular dynamics (AIMD), or Rattled relaxation. SNUMAT provides API access to its database of around 10,000 synthesized materials and their DFT properties.
Other datasets, such as \textbf{Open Catalyst 2020 (OC20)}~\cite{chanussot2021open} and \textbf{MPTrj}~\cite{deng2023chgnet}, have been designed with large-scale adsorption energies and structures on catalyst surfaces, interatomic potentials, and atomic trajectory. For exploring molecular property, \textbf{Alexandria} dataset \cite{schmidt2024improving} offers an open collection of approximately 5 million DFT-calculated molecular structures. The dataset is categorized by material type, including 3D, 2D, and 1D materials. \textbf{QM9}~\cite{ruddigkeit2012enumeration,ramakrishnan2014quantum} includes over 100,000 small organic molecules with DFT-calculated quantum chemical properties. To support molecular design, \textbf{Guacamol} \cite{brown2019guacamol} and \textbf{Moses} \cite{polykovskiy2020molecular} are benchmarks for evaluating models trained for molecular generation task. Some chemical datasets can also serve as additional resources for retrieving specific information about materials. For example, 
\textbf{PubChem} \cite{kim2025pubchem} is the largest open database of chemical information including name, formula, structure, and other identifiers. The database hosts around 121M chemical structures and hundreds of thousands of proteins and genes. \textbf{ZINC} \cite{irwin2020zinc20, sterling2015zinc} is another chemical database of compounds for virtual screening, including over one billion of molecules.
\textbf{ChEMBL}~\cite{zdrazil2024chembl, davies2015chembl} is a database for drug discovery, including bioactive molecules and drug-like properties.  

Despite this progress, the landscape remains fragmented. Table~\ref{tab:coverage} provides a cross-sectional view of widely used computational and experimental datasets in materials science with respect to the types of materials they support and the physical length scales they span. While most datasets are heavily skewed toward inorganic and atomically resolved systems, a few, particularly NOMAD and ChEMBL extend into organic, polymeric, and biomaterial domains. This uneven distribution highlights a critical gap in the availability of structured, large-scale data for polymers, composites, and mesoscale to macroscale phenomena. The dominance of atomistic simulation datasets has naturally shaped the focus of existing foundation models, many of which are optimized for crystalline inorganic compounds. As the field moves toward more generalizable and multimodal models, addressing this data imbalance is essential for expanding foundation model applicability across underexplored material classes and real-world scales. This challenge underscores the need for systematic data curation and benchmarking efforts that extend beyond atomic configurations, enabling models to reason across structure, property, processing, and scale.  Moreover, few datasets provide truly multimodal, property-aligned examples—such as atomic structure paired with synthesis route and experimental spectra—thereby limiting the development of generalist models capable of reasoning across diverse data types. Many datasets also suffer from compositional, elemental, and phase-type biases, with overrepresentation of stable oxides and underrepresentation of disordered or metastable materials~\cite{yang2024mattersim}. Ongoing efforts in data augmentation, transfer learning, and active learning aim to address these limitations by prioritizing data diversity, especially in low-resource domains like polymers, amorphous systems, and biomaterials \cite{choi2025perspective, lee2023towards}. 

\begin{table}[ht]
\centering
\scriptsize
\caption{Coverage of computational and experimental datasets across material types and length scales. Checkmarks indicate areas where the dataset or model is primarily applied or relevant.}
\label{tab:coverage}
\begin{tabular}{lcccc|ccccc}
\toprule
& \multicolumn{4}{c}{Material Type} & \multicolumn{4}{c}{Length Scale} \\
\midrule
\textbf{Dataset / Model} & \textbf{Inorganic} & \textbf{Organic} & \textbf{Polymers} & \textbf{Bio} & \textbf{Atomic} & \textbf{Nano} & \textbf{Meso} & \textbf{Macro} \\
& & & & & $< 10^{-9} m$ & $10^{-9}-10^{-6} m$ & $10^{-6}-10^{-3} m$ & $> 10^{-3} m$ \\
\midrule
\multicolumn{9}{l}{\textbf{Datasets}} \\
\midrule

ICSD \cite{zagorac2019recent} & \cmark & \cmark & & & \cmark & \cmark & & \\
Materials Project (MP) \cite{jain2013commentary} & \cmark & & & & \cmark & \cmark & & \\
OQMD \cite{kirklin2015open,saal2013materials} & \cmark & & & & \cmark & & & \\
NOMAD \cite{draxl2019nomad} & \cmark & \cmark & \cmark & \cmark & \cmark & \cmark & & \\
OMat24 \cite{barroso2024open} & \cmark & & & & \cmark & \cmark & & \\
SNUMAT \cite{snumat} & \cmark & & & & \cmark & \cmark & & \\
MPTrj \cite{deng2023chgnet} & \cmark & & & & \cmark & \cmark & & \\
Open Catalyst 2020 (OC20) \cite{chanussot2021open} & \cmark & & & & \cmark & \cmark & & \\
Alexandria \cite{schmidt2024improving} & \cmark & & & & \cmark & & & \\
QM9 \cite{ruddigkeit2012enumeration,ramakrishnan2014quantum} & & \cmark & & & \cmark & & & \\
Guacamol \cite{brown2019guacamol} & \cmark & \cmark & & & \cmark & \cmark & & \\
Moses \cite{polykovskiy2020molecular} & & \cmark & & & \cmark & \cmark & & \\
ZINC \cite{sterling2015zinc,irwin2020zinc20} & & \cmark & & & \cmark & \cmark & & \\
ChEMBL \cite{zdrazil2024chembl, davies2015chembl} & & \cmark & & \cmark & \cmark & \cmark & & \\
\bottomrule
\end{tabular}
\end{table}

\subsubsection{LLM development datasets}
With the rapid development of LLMs such as ChatGPT, Gemini, Llama, or Claude, more real-world AI-assisted systems have been built to bring useful domain-specific applications to users. The scientific domain, especially materials science, is an important goal in achieving artificial general intelligence. Developing LLMs requires several stages including training, instruction fine-tuning, and evaluation, which necessitate several large datasets for various tasks. 

For training LLMs with general knowledge, \textbf{RedPajama} \cite{weber2024redpajama}, \textbf{BookCorpus} \cite{zhu2015aligning}, \textbf{Wikitext} \cite{merity2016pointer} and \textbf{Common Crawl} \cite{ccrawl} are common open-source datasets including billions of entries for LLM pretraining. Other datasets of literature papers and patterns are also used for training LLMs in scientific domains. \textbf{MatScholar} \cite{tshitoyan2019unsupervised,weston2019named} includes over five million paper abstracts in the materials science domain. \textbf{SciDocs}~\cite{cohan2020specter} offers a foundation for language model pretraining, focusing on scientific documents. \textbf{USPTO} \cite{uspto} offers Open Data Portal (ODP) for accessing US patterns, which enables researchers to discover and extract related information to specific domains. In addition to chemical compounds, \textbf{PubChem} \cite{kim2025pubchem} also provides 53M patents and 42M research papers. All these literature-related datasets can sufficiently enable the training of LLMs for materials science. \textbf{SciCap} \cite{hsu2021scicap} is a multimodal dataset with scientific figure-caption pairs extracted from over 290,000 arXiv papers, which is suitable for training multimodal models like vision-language models. Comprehensive knowledge bases are also essential to operate LLM-based agentic systems as they require a sufficient amount of information to generate hypotheses, do planning, or provide accurate answers. To build LLM-based agents for materials science, \textbf{MatSciKB} \cite{zhang2024honeycomb}, which is part of the HoneyComb work, includes diverse sources of data related to materials science—such as arXiv papers, Wikipedia pages, textbooks, multiple-choice datasets, formulas, and GPT-generated entries—to support the task of knowledge retrieval, hypotheses generation, and experiment planning. 

Instruction fine-tuning (IFT) is another critical step when developing LLM-based chat assistants, enhancing the capability of generating human-like conversations. Several datasets are introduced to support IFT including general Q\&A like \textbf{OpenOrca} \cite{mukherjee2023orca} and \textbf{WebInstructSub} \cite{yue2024mammoth2}, and science-related Q\&A extracted from science exams or knowledge bases of different subjects, such as \textbf{MathQA} \cite{amini2019mathqa}, \textbf{MatbookQA} \cite{mishra2024foundational}, \textbf{MaScQA} \cite{zaki2024mascqa}, \textbf{ARC} \cite{clark2018think}, \textbf{PIQA} \cite{bisk2020piqa}, \textbf{SciQ} \cite{welbl2017crowdsourcing} and \textbf{ScienceQA} \cite{lu2022learn}. \textbf{MatSci-Instruct} \cite{song2023honeybee}, which is proposed in HoneyBee work, is an attempt to build an instruction fine-tuning dataset specified for materials science with around 52k instructions. \textbf{MatSci community disclosure}\footnote{MatSci Com. Dis., \url{https://matsci.org/}} is a forum for discussions of topics related to materials science, which is also a helpful Q\&A data source. 

While several LLMs have been introduced for materials science, the cost of deploying an LLM on local machines is expensive due to the high cost of running GPU servers. Cloud-based pretrained LLMs, such as ChatGPT and Gemini, are still cost-efficient options for materials science researchers. Hence, the task of evaluating and benchmarking pretrained LLMs in the materials science domain has been catching the attention of the research community to assess the performance of pretrained models on downstream tasks related to materials science. To evaluate language models on materials science text, \textbf{MatSciNLP} \cite{song2023matsci} provides a benchmark comprising seven tasks, including Named Entity Recognition (NER), relation, sentence, and paragraph classification, event argument extraction, synthesis action retrieval, and slot filling.
\textbf{LLM4Mat-Bench} \cite{niyongabo2024llm4mat} offers a benchmark specifically focused on evaluating LLM capabilities in property prediction, particularly for crystalline materials. It includes around 1.9M of structures formed by 10 public materials datasets with 45 distinct properties. Focusing on multimodalities, LLM4Mat-Bench covers several modalities, such as crystal compositions, CIFs, and corresponding textual descriptions. Several models are evaluated to show their performance in material property prediction tasks, such as CGCNN, MatBERT, LLM-Prop, Llama, Gemma, and Mistral. 
\textbf{LLM4Mol} \cite{zhong2024benchmarking} introduces a benchmark aimed at evaluating the ability of LLMs to perform molecular prediction tasks. The benchmark covers six molecule datasets in two important prediction tasks: classification and regression. Prompt engineering is utilized to instruct LLMs, such as GPT-family and Llama-family models, to predict outcomes using both zero-shot and few-shot prompting techniques. 
\textbf{MACBENCH} \cite{alampara2024macbench} is a multimodal benchmark comprising 628 questions designed to evaluate the multimodal reasoning capabilities of AI models in materials science and chemistry tasks. MACBENCH encompasses tasks of fundamental scientific understanding, data extraction from visual information, and practical knowledge.  Its diverse visual inputs include laboratory images, band structures, crystal structures, and atomic force microscopy images paired with multiple-choice questions. This benchmark addresses the multimodal nature of materials science and captures the tacit knowledge and laboratory skills.

While the descent effort of introducing training and benchmarking datasets for materials science has been introduced, the need for comprehensive and multi-task datasets still remains. Furthermore, there is also a lack of multimodal datasets for training multimodal LLM.

\subsection{Tools and Infrastructures}
In this section, we review useful tools and infrastructures designed for the analysis and management of materials data, as well as for the development of machine learning models. A summary of key resources is provided in Table~\ref{tab:resources}.

\begin{table}[ht]
\centering
\scriptsize
\caption{Key tools and infrastructures supporting foundation models in materials science.}
\label{tab:resources}
\begin{tabular}{p{1in}p{1.2in}p{1in}p{1.4in}p{0.55in}}
\toprule
\textbf{Name} & \textbf{Primary Functionality} & \textbf{Supported Modalities} & \textbf{Supported Models/Applications} & \textbf{Access} \\
\midrule
\multicolumn{5}{c}{\textbf{Materials data analysis and management tools}} \\\midrule
Materials Project API~\cite{jain2013commentary} & Access to millions of computed materials entries for downstream use & Structures, formation energy, bandgap & Various via API & Open-source \\
OPTIMADE~\cite{andersen2021optimade} & Access to millions of material structures and information & Structures, DFT calculations, references & Various via API, Python tools & Open-source \\

MPTrj \cite{deng2023chgnet} & Trajectory-level data for training universal force fields & Forces, atomic positions, energies & MACE-MP-0, NequIP, Allegro & Open-source \\
Pymatgen~\cite{ong2013python} & Tool for processing, analyzing, and visualizing material data & Crystal structures, phase diagrams, properties & Various & Open-source \\
M$^2$Hub \cite{du2023m} & Tool for data processing, model implementation, and training & Various & Various & Open-source \\
FireWorks \cite{jain2015fireworks} & Tool for end-to-end scientific workflows & Various & Job monitor, databases, high-performance computing systems & Open-source \\
Maggma \cite{maggma} & Tool for data queries, transformations, and storage & Various & Databases & Open-source \\
ToolHub \cite{zhang2024honeycomb} & Unified interface with several tools to access online sources & Text & HoneyComb & Open-source \\

\midrule
\multicolumn{5}{c}{\textbf{Model development tools}} \\\midrule
MatBench~\cite{dunn2020benchmarking} & Benchmarking model performance across 13 materials tasks & Structure, composition, scalar properties & ALIGNN, MEGNet, Roost, CGCNN & Open-source \\
ALIGNN-FF (Toolkit)~\cite{choudhary2023unified} & Training force fields from trajectories using message passing & Forces, energies, atomic trajectories & ALIGNN, ALIGNN-FF & Open-source \\
FORGE~\cite{yin2023forge} & Pretraining and fine-tuning for foundation models & Graphs, atomic structures, text & ChemGPT, ALIGNN, MoLFormer variants & Open-source \\
OC20 (Tasks)~\cite{chanussot2021open} & Large-scale catalyst surface dataset and ML challenge suite & Adsorbates, surfaces, trajectories, energies & DimeNet++, SchNet, GemNet-OC & Open-source \\
Open MatSci ML Toolkit~\cite{miret2022open} & Standardized access to datasets, training and benchmarking & Structure, composition, text & ALIGNN, CGCNN, MEGNet & Open-source \\
A-Lab (Autonomous Lab)~\cite{szymanski2023autonomous} & Robotic synthesis and closed-loop materials optimization & Protocols, text, synthesis metadata & Surrogate models, active learners & Closed-lab \\
GNoME Infrastructure (Discovery Engine)~\cite{merchant2023scaling} & Active learning and scalable DFT validation for discovery & Graphs, formation energy, symmetry & GNN ensemble, GNoME & Internal \\
MatterSim Infrastructure (HPC Platform)~\cite{yang2024mattersim} & Large-scale pretraining on 17M DFT-labeled structures & Atomic structures, forces, temperature & MatterSim & Internal \\
Ollama \cite{ollama} & LLM deployment tool & Text & Various LLMs & Open-source \\
LangChain \cite{langchain} & LLM-driven application development tool & Various & LLMs, databases, web scrapping & Open-source \\

\bottomrule
\end{tabular}
\end{table}

\subsubsection{Materials data analysis and management tools}
In the effort to enhance materials data analysis and management, several packages and libraries are developed to support materials science research. \textbf{Pymatgen} (Python Materials Genomics) \cite{ong2013python} is an open-source Python library for processing, analyzing, and visualizing crystal structures, phase diagrams, and material properties in different formats (e.g., VASP, ABINIT, CIF, XYZ). A Python-based web app framework, named \textbf{Crystal Toolkit} \cite{horton2023crystal}, is introduced to provide an interactive interface for exploring materials science information. \textbf{M$^2$Hub} \cite{du2023m} is another Python-based toolkit that provides machine learning practice for materials science. It supports users with the whole pipeline creation, covering from data processing tasks (downloading and preprocessing) to model implementation and training. Several additional Python packages have been developed to design and manage end-to-end scientific workflows. These include \textbf{FireWorks} \cite{jain2015fireworks}, \textbf{Custodian} \cite{ong2013python}, \textbf{Atomate} \cite{mathew2017atomate}, and \textbf{Jobflow} \cite{rosen2024jobflow}, which facilitate tasks such as job writing, execution, automation, management, and result analysis. For the purpose of building a data pipeline and database, \textbf{Emmet} \cite{emmet} and \textbf{Maggma} \cite{maggma} are developed by Materials Project to support users with data queries, data transformations, and data storage. HoneyComb provides \textbf{ToolHub}, a unified interface consisting of several tools for accessing online sources such as Google search, Wikipedia search, or materials analysis tools. To build an API for exchanging material-related data, \textbf{OPTIMADE} \cite{andersen2021optimade} is an open-source tool for providing access to approximately 59M structures from 29 databases. 

\subsubsection{Model development tools}
On the infrastructure and tooling front, several modular and open-source tools have emerged to support training, fine-tuning, and evaluation of both deep learning and foundation models. Focusing on the evaluation of machine learning and deep learning models trained for materials science, \textbf{MatBench}~\cite{dunn2020benchmarking} leverages materials data from the aforementioned datasets, like MP, to formulate a comprehensive set of benchmarking tasks. \textbf{FORGE}~\cite{yin2023forge} provides a flexible pretraining and benchmarking framework for large-scale scientific models, supporting graph-based architectures and self-supervised objectives across chemistry and materials datasets. Autonomous laboratories such as \textbf{A-Lab}~\cite{szymanski2023autonomous} further illustrate the integration of foundation models with experimental loops, enabling robotic synthesis, online decision-making, and self-improving discovery. \textbf{Open MatSci ML Toolkit}~\cite{miret2022open} offers a curated interface to datasets, model architectures, and evaluation protocols, aimed at improving reproducibility and benchmarking across tasks. Toolkits like \textbf{ALIGNN-FF}~\cite{choudhary2023unified} extend message-passing neural networks to force field learning, with a focus on universal atomistic simulation capabilities. 
\textbf{Geom3D} is a platform for geometric modeling on 3D structures and integrates MatBench and QMOF datasets and several geometry graphical neural networks. \textbf{JARVIS-Leaderboard} is an open-source and community-driven platform and allows users to set up benchmarks with various types of input data and different tasks. \textbf{GT4SD} \cite{manica2023accelerating} is a Python toolkit for developing and training generative models for scientific discovery. Training large-scale models also necessitates extensive infrastructures or resources. For example, \textbf{MatterSim}~\cite{yang2024mattersim} is trained on over 17M DFT-labeled structures using industrial-scale infrastructures, while \textbf{GNoME}~\cite{merchant2023scaling} leverages active learning in combination with distributed DFT validation. These efforts highlight the importance of scalable computational resources and data workflows in developing large foundation models.

The development of LLM and LLM-based agents requires the integration of LLM deployment, environment interaction, and external resource access. Several toolkits have been introduced to support these tasks. For instance,  \textbf{Ollama} \cite{ollama} provides a platform to deploy LLMs in local machines, covering a diverse set of open-source pretrained models. \textbf{LangChain} \cite{langchain} is a framework for developing LLM-based applications, such as Retrieval Augmented Generation (RAG) or LLM agents. This framework provides the flexibility to connect different components and integrate third-party applications into the unified AI system. Similarly, \textbf{AutoGen} \cite{autogen} and \textbf{CrewAI} \cite{crewai} are popular frameworks for building role-playing LLM agents, facilitating the development of multi-agent systems. To support the deployment of an LLM-based agentic system which connects LLM to external data resources, \textbf{LlamaIndex} \cite{llamaindex} is specifically introduced to provide the solution for this task.

Despite the scale of these efforts, access remains uneven. Most large-scale pretraining runs rely on proprietary or institutional resources, limiting reproducibility and broader community participation. Continued development of open infrastructures, shared checkpoints, and model zoos—alongside high-quality dataset curation—will be essential to democratize foundation model development and foster inclusive progress across materials science. Beyond static tooling, some models integrate tightly with experimental and high-performance computing pipelines.


\section{Successes, Limitations, Challenges, and Future}
\label{sec:successes_limitations_future}

\begin{figure}[ht]
    \centering
    \includegraphics[width=.95\linewidth]{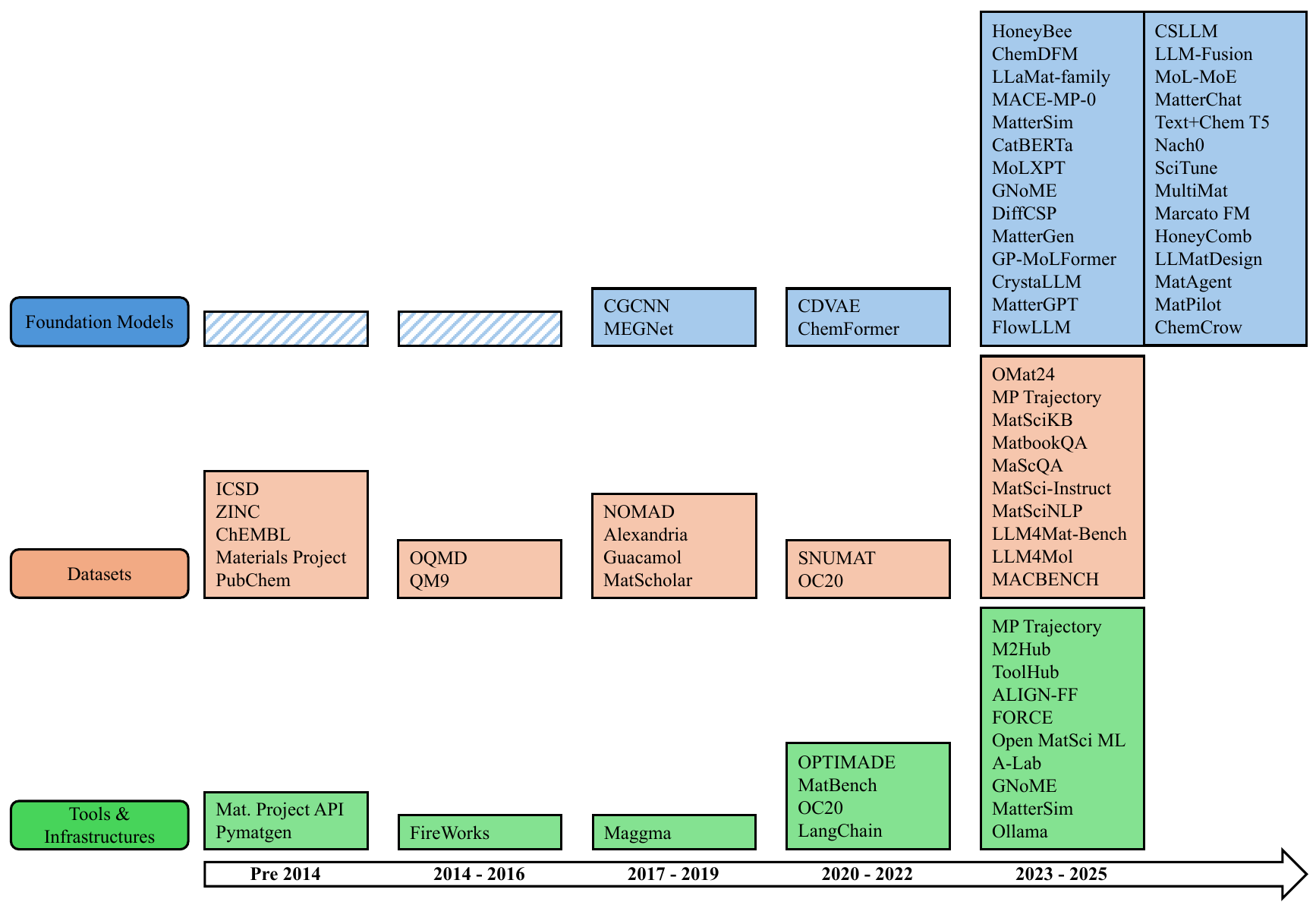}
    \caption{Development of AI in materials science over time: foundation models, datasets, and tools and infrastructure.}
    \label{fig:summary_through_time}
\end{figure}

With the recent rapid development of Artificial Intelligence, its application to scientific domains such as materials science is inevitable. AI offers predictive capabilities, insightful analysis, scalability, and efficiency that can significantly advance materials research. While the materials science community has been developing and utilizing datasets and programming toolkits for decades, foundation models are a more recent addition. Figure~\ref{fig:summary_through_time}, which charts the release years of foundation models, datasets, and infrastructure tools, illustrates the accelerating convergence of AI and materials science. In this section, we first review the early successes of AI in the field, then examine current limitations and challenges, and finally discuss promising future research directions.

\subsection{Early Successes}

The growing maturity of foundation models in materials science is reflected in a series of high-impact applications that demonstrate their ability to augment or outperform traditional computational and experimental techniques. These successes span materials discovery, molecular dynamics, inverse design, autonomous synthesis, and literature-based information extraction, catalyzing a shift toward generalist, scalable, and efficient AI systems for scientific research \cite{merchant2023scaling,yang2024mattersim,zeni2023mattergen,livne2024nach0,szymanski2023autonomous}.

One of the most significant achievements is the large-scale discovery of new materials. The GNoME (Graph Networks for Materials Exploration) system exemplifies this potential by using ensembles of graph neural networks trained on DFT formation energy data, symmetry-aware crystal generation, and iterative DFT validation in an active learning loop. GNoME identified over 2.2M new stable inorganic materials—an order-of-magnitude leap beyond prior efforts—and discovered over 45,000 new crystal prototypes occupying previously unexplored regions of chemical space~\cite{merchant2023scaling}. Meanwhile, MatterSim, a universal machine-learned interatomic potential trained on over 17M DFT-labeled configurations, enables zero-shot molecular dynamics simulations with energy prediction errors below 50 meV/atom, even in high-temperature and non-equilibrium regimes. Other models such as MACE-MP-0 have demonstrated near-DFT accuracy using message-passing neural networks applied to periodic systems. 

In generative modeling and inverse design, diffusion- and language-based foundation models are enabling property-guided exploration of materials space. MatterGen combines unsupervised pretraining with adapter-based fine-tuning to generate chemically valid, structurally stable, and property-optimized inorganic crystals. It outperforms prior models in diversity, novelty, and closeness to DFT-derived properties. Language-based models like GP-MoLFormer extend molecular design to functional materials and drug-like compounds through property-conditioned generation. 

On the text mining and scientific reasoning front, models like nach0 and ChemDFM demonstrate cross-modal capabilities for molecule generation, retrosynthesis, property prediction, and question answering. Trained on structured and unstructured chemistry corpora, these models extract synthesis routes, interpret SMILES strings, and enable high-throughput knowledge extraction from literature.

Although LLM-based agents are still in their early stages and continue to evolve toward more sophisticated and practical applications, recent developments—such as HoneyComb, LLMatDesign, MatAgent, and MatPilot—demonstrate promising progress in building agentic systems for materials science. Beyond prediction and generation, several of these works incorporate foundation models into agentic workflows that support planning, reasoning, and tool integration. These systems offer important conceptual frameworks that aid researchers and practitioners in understanding, analyzing, and automating complex tasks more efficiently. For example, MatAgent, LLMatDesign, and MatPilot utilize LLMs to coordinate multi-step processes such as candidate screening, simulation planning, and synthesis route generation, all with minimal human intervention.

Foundation models have also been deployed in autonomous systems: the A-Lab integrates language planning, thermodynamic reasoning, robotics, and active learning into a closed-loop experimental platform. Operating over 17 days, A-Lab synthesized 41 of 58 targeted materials with a 71\% success rate and minimal human intervention.

\subsection{Limitations and Challenges}

Despite recent promising developments, foundation models in materials science remain an emerging technology with substantial limitations. One of the most critical challenges is the accurate modeling of long-range interactions such as electrostatics, dispersion, and magnetism. Local message-passing architectures like those in MatterSim and MACE-MP-0 effectively capture short-range bonding but fall short for systems dominated by nonlocal physics~\cite{batatia2023foundation, yang2024mattersim}. Although empirical corrections can partially mitigate this issue, comprehensive and efficient solutions remain elusive.

Generalizability presents a significant challenge in the training of foundation models for materials science. A core objective in this field is the discovery of novel materials with unprecedented properties tailored for specific applications. Foundation models trained on existing material distributions should demonstrate the capability to be adaptive to out-of-distribution materials or unseen domains. Moreover, materials behave differently under different conditions, such as super high/low temperatures or pressure. Vision-based foundation models, used to detect defects from images or videos, may also struggle as the visual appearance of materials evolves over time. In addition, many underrepresented material classes—such as polymers, disordered solids, and biomaterials—remain challenging due to sparse, noisy, or long-tail data distributions, which limit the generalizability of current FMs. These challenges underscore the need for more in-depth research into the design and training of robust foundation models in materials science.

Integration with experimental workflows remains limited. Most foundation models are not trained with synthesizability, phase stability, or safety in mind. Although MatterGen can generate candidate materials optimized for electronic or mechanical properties, it cannot guarantee experimental feasibility~\cite{zeni2023mattergen}. Few models jointly optimize for both computational performance and laboratory viability. Similarly, multimodal integration remains underdeveloped. While models like nach0 and ATLANTIC explore text-structure-property fusion, most foundation models operate in unimodal spaces~\cite{livne2024nach0, munikoti2023atlantic}, lacking the ability to simultaneously reason over images, spectra, simulation outputs, and experimental metadata.

Interpretability is another pressing concern. As these models increase in size and complexity, understanding their internal representations and assessing their trustworthiness becomes increasingly difficult. Generative models like nach0 and MatterGen can produce physically implausible results despite being syntactically valid~\cite{livne2024nach0, zeni2023mattergen}. Black-box behavior poses risks in high-stakes applications like battery chemistry or catalysis. Furthermore, foundation models are often evaluated on in-distribution test sets, whereas materials discovery requires extrapolation to new chemistries, properties, or structural motifs. While models like GNoME have demonstrated generalization across broad chemical spaces, the reliability of predictions in entirely novel regimes is unclear~\cite{merchant2023scaling}.

Data bias is another significant limitation. Current training corpora overrepresent stable, inorganic, equilibrium-phase systems—especially oxides—while underrepresenting high-entropy alloys, amorphous phases, soft materials, and molecular crystals. While active learning and off-equilibrium sampling strategies aim to address this, coverage remains incomplete. This imbalance reduces the generality of models and limits their utility in less-studied or industrially relevant materials domains.

Several challenges arise in the design and deployment of LLMs and LLM agents, necessitating further research. LLMs rely heavily on the availability of data to train the language models. The data collection and preprocessing stages are critical for ensuring the quality of the generated content, particularly when natural and human-like responses are expected. These stages often require significant resources and costs to curate high-quality training data. 

Another key challenge is LLM hallucination, especially when models are applied to unfamiliar or out-of-distribution scenarios. Such issues may stem from corrupted training data, limitations in model architecture, or ambiguous user instructions. This might arise due to the corrupted training data, limited model design, or ambiguous user instructions. Finally, biosafety and chemical safety are highly important while working with materials science experiments. These models may generate syntactically valid but physically implausible results, or propose unsafe synthesis conditions without proper grounding. Experimental plans generated by LLM agents might offer excellent opportunities to analyze and discover new findings but also contain potential risks if not thoroughly reviewed by human experts. These considerations highlight the urgent need for comprehensive evaluation benchmarks and training frameworks (e.g. supervised fine-tuning, reinforcement learning, human-in-the-loop) tailored to specific tasks in materials science, to ensure the safe and effective deployment of LLM and LLM agents.

Supporting the development of autonomous, LLM-based agentic systems also requires the creation of open-source, highly scalable, and integrable materials science tools and infrastructure. High-quality, standardized tools can significantly reduce the time and resources required for implementation, thereby accelerating the transition of agentic systems from research prototypes to practical, real-world applications accessible to researchers and end users. 

The substantial compute requirements for training and deploying foundation models raise significant concerns about accessibility. For example, models such as MatterSim and GNoME demand tens of thousands of GPU hours, while large language models like GPT, Gemini, and LLaMA require millions of GPU hours, often relying on proprietary computing resources—placing such efforts beyond the reach of most academic researchers. Open-source tools and frameworks like FORGE, Open MatSci ML, and ALIGNN-FF represent promising steps toward democratizing access and enabling smaller-scale experimentation. Nevertheless, despite these advances, fostering broader participation and ensuring sustainable development across the research community remain ongoing challenges that must be addressed to fully realize the potential of foundation models in materials science.

\subsection{Future Direction}

To address these limitations and chart the path forward, several directions merit emphasis. Future foundation models must incorporate stronger physical laws to align with quantum mechanics, thermodynamics, and symmetry principles. Architectures with built-in equivariance or physics-informed constraints may improve both accuracy and interpretability. Cross-modal models that integrate crystal structures, synthesis routes, spectra, images, and text hold the key to more holistic materials reasoning. Expanding training datasets to include failed experiments, synthesis outcomes, and non-traditional materials classes will improve generality and real-world relevance. Another direction is the principled development of foundation models for materials science, which are typically data hungry, that can assimilate both low-fidelity data (cheap to collect) and high-fidelity data (expensive to collect) to circumvent the high cost of building a large high-fidelity training dataset and mitigate data sparsity concerns. Progress will also depend on aligning text, structure, spectra, and image-based data through high-quality, co-registered datasets and multimodal pretraining strategies. Unified tokenization schemes and modality adapters may be necessary to bridge diverse formats.

Active and continual learning will be essential for deploying foundation models in dynamic, exploratory settings. Integration with robotic labs, uncertainty-aware retraining, and human-in-the-loop frameworks will allow models to learn from failure, adapt to new tasks, and serve as collaborators in discovery pipelines. Efficient training paradigms—modular, sparse, or low-rank—will be necessary to reduce the environmental and financial cost of model development. The growing potential for combining large language models like GPT-4 with specialized materials tools signals a future where AI-driven autonomous agents could revolutionize materials science research, making these models indispensable to materials discovery. Alongside technical progress, the field must prioritize open benchmarks, reproducible evaluation, interpretability tooling, and shared infrastructures to ensure equitable progress.

Data governance and data curation are essential steps in enhancing the quality of training data. Human feedback data is also a crucial component in achieving more natural conversations between users and models. The principles of trustworthy AI can be employed to develop models that prioritize privacy, safety, and reliability. Techniques such as adversarial learning and machine unlearning contribute to these goals by improving model robustness and enabling the removal of sensitive or erroneous data. Finally, integrating human-in-the-loop approaches offers a promising strategy for incorporating human oversight during task execution, thereby improving both accuracy and accountability.

Trustworthy AI principles such as uncertainty quantification, adversarial testing, machine unlearning, and human feedback must guide FM deployment in scientific domains. Ensuring safe, robust, and interpretable behavior will be essential for adoption in high-stakes applications. Ultimately, the success of foundation models in materials science hinges on interdisciplinary collaboration, standardized data practices, and deep integration of domain knowledge and experimental feedback. With these components in place, foundation models have the potential to become not just predictive engines, but collaborative agents for reasoning, creativity, and exploration across materials science.


\section{Conclusion}
\label{sec:conclusion}

The emergence of foundation models represents a profound shift in materials science, echoing the transformations currently underway in fields like natural language processing, computer vision, and molecular biology. These large, pretrained, and general-purpose models have demonstrated remarkable capabilities in property prediction, atomistic simulation, materials generation, and language-based scientific reasoning. What distinguishes foundation models from previous generations of machine learning tools is not only their scale, but also their transferability, emergent behavior, and versatility across tasks and data modalities.

In this survey, we propose a clear taxonomy of materials foundation models, organizing them according to tasks—including data extraction, atomistic simulation, property prediction, materials design and discovery, process optimization, and multiscale modeling—and by model type, ranging from specialized unimodal foundation models to ambitious multimodal foundation models, as well as emerging autonomous and flexible LLM agents. Within these categories, unimodal foundation models such as GNoME, MatterSim, MACE-MP-0, and MatterGen collectively demonstrate the feasibility of AI-driven materials discovery at unprecedented scales and speeds. Multimodal foundation models like MatterChat, Text+Chem T5, and nach0 exemplify efforts to integrate multimodal materials information, thereby enhancing the quality of outcomes. The recent emergence of LLM-powered agents (e.g., HoneyComb, MatAgent, LLMatDesign) highlights a shift toward systems that not only analyze data but also reason, plan, and interact across formats such as text, structure, images, and protocols.

While these successes are undeniably impressive, our review also surfaces significant limitations and open challenges. Issues of long-range interaction modeling, data imbalance, interpretability, and generalization under extrapolation remain unresolved. Similarly, the computational and infrastructural demands of training large foundation models limit their accessibility and reproducibility in the broader materials research community. Integrating these models into experimental workflows and achieving multimodal reasoning capabilities remain aspirational goals for the near term.

Looking ahead, the field is clearly moving toward scalable, multimodal, and human-AI collaborative discovery systems. Foundation models that can reason over atomic structures, properties, synthesis procedures, and scientific text within a unified framework have the potential to dramatically accelerate materials research, while also opening new frontiers in autonomous laboratories, hypothesis-driven design, and interactive scientific exploration. Realizing this vision will require not only continued advances in model architectures, datasets, and computational infrastructures, but also a cultural shift toward openness, interdisciplinarity, and rigorous, transparent evaluation. To realize their full potential, foundation models must be not only accurate and scalable, but also trustworthy, interpretable, and safe. Community efforts toward reproducibility, human-in-the-loop systems, governance, and equitable access will be critical in guiding their responsible deployment.

Ultimately, foundation models offer a powerful new paradigm for materials science; one that complements and enhances, rather than replaces, human expertise. The challenge for the research community now is to harness these tools responsibly, inclusively, and creatively, ensuring that the benefits of AI-driven materials discovery are both scientifically rigorous and broadly accessible. Foundation models are poised to become collaborative scientific agents, transforming how we explore, reason about, and design the materials of the future.

\section*{Acknowledgement}
This work was supported in part by the National Institute of General Medical Sciences of National Institutes of Health under award P20GM139768, and the Arkansas Integrative Metabolic Research Center at the University of Arkansas.

\bibliographystyle{unsrtnat}

\end{document}